\documentclass[journal,12pt,onecolumn]{IEEEtran}
\ifCLASSINFOpdf
\else
\fi

\usepackage{bm}
\usepackage{CJK}
\usepackage{indentfirst}
\usepackage{amsmath}
\usepackage{multirow}
\usepackage{threeparttable}
\usepackage{tabularx}
\usepackage{graphicx}
\usepackage{subfigure}
\usepackage{float}
\usepackage{color}
\usepackage{booktabs}
\usepackage{amsmath,amssymb}
\usepackage{epstopdf}
\usepackage{inputenc}
\usepackage{diagbox}
\usepackage{cite}
\makeatletter
\newif\if@restonecol
\makeatother

\usepackage[linesnumbered,ruled,vlined]{algorithm2e}   
\usepackage{algpseudocode}
\usepackage{amsmath}

\begin{document}
\title{Slow-varying Dynamics Assisted \\
Temporal Capsule Network for \\
Machinery Remaining Useful Life Estimation}
\author{Yan~Qin,~\IEEEmembership{Member, IEEE,}
             Chau~Yuen,~\IEEEmembership{Fellow,~IEEE,}
             Yimin~Shao,~\IEEEmembership{}
             Bo~Qin,~\IEEEmembership{}
             and Xiaoli~Li,~\IEEEmembership{Senior Member, IEEE}
\thanks{This work is supported in part by the A*STAR-NTU-SUTD Joint Research Grant on Artificial Intelligence Partnership under Grant RGANS1906, in part by the Singapore Ministry of National Development and the National Research Foundation, Prime Minister’s Office under the Cities of Tomorrow (CoT) Research Programme (Award No. COT-V2-2021-1), and in part by National Natural Science Foundation of China under Grant 61903327. Any opinion, findings, and conclusions or recommendations expressed in this material are those of the author(s) and do not reflect the views of the Singapore Ministry of National Development and National Research Foundation, Prime Minister’s Office, Singapore. (Corresponding author: Chau Yuen.)}
\thanks{Y. Qin and C. Yuen are with the Engineering Product Development Pillar, Singapore University of Technology and Design, 8 Somapah Road, 487372 Singapore. (e-mail: yan.qin@ntu.edu.sg, yuenchau@sutd.edu.sg)}
\thanks{Y. Shao is with the State Key Laboratory of Mechanical Transmissions, Chongqing University, 400044 Chongqing, P. R. China. (e-mail: ymshao@cqu.edu.cn)}
\thanks{B. Qin is with the School of Mechanical Engineering, Inner Mongolia University of Science and Technology, 014010 Baotou, P. R. China. (e-mail: boqin@imust.edu.cn)}
\thanks{X. Li is with the Institute for Infocomm Research (I2R), A*STAR, 138632 Singapore. (e-mail: xlli@i2r.a-star.edu.sg)}
}

%

\markboth{}
{Shell \MakeLowercase{\textit{et al.}}: Bare Demo of IEEEtran.cls for IEEE Journals}
%

\maketitle
\begin{abstract}
Capsule network (CapsNet) acts as a promising alternative to the typical convolutional neural network, which is the dominant network to develop the remaining useful life (RUL) estimation models for mechanical equipment. Although CapsNet comes with impressive ability to represent the entities' hierarchical relationships through a high-dimensional vector embedding, it fails to capture long-term temporal correlation of run-to-failure time series measured from degraded mechanical equipment. On the other hand, the slow-varying dynamics, which reveals the low-frequency information hidden in mechanical dynamical behaviour, is overlooked in the existing RUL estimation models, limiting the utmost ability of advanced networks. To address the aforementioned concerns, we propose a Slow-varying Dynamics assisted Temporal CapsNet (SD-TemCapsNet) to simultaneously learn the slow-varying dynamics and temporal dynamics from measurements for accurate RUL estimation. First, in light of the sensitivity of fault evolution, slow-varying features are decomposed from normal raw data to convey the low-frequency components corresponding to the system dynamics. Next, the long short-term memory (LSTM) mechanism is introduced into CapsNet to capture the temporal correlation of time series. To this end, experiments conducted on an aircraft engine and a milling machine verify that the proposed SD-TemCapsNet outperforms the mainstream methods. In comparison with CapsNet, the estimation accuracy of the aircraft engine with four different scenarios has been improved by 10.17$\%$, 24.97$\%$, 3.25$\%$, and 13.03$\%$ concerning the index root mean squared error, respectively. Similarly, the estimation accuracy of the milling machine has been improved by 23.57$\%$ compared to LSTM and 19.54$\%$ compared to CapsNet.
\end{abstract}

\section{Introduction}
\IEEEPARstart{M}{odern} mechanical systems are ever-increasingly revolutionized by the rapid development of the industrial Internet of Things and advanced sensor technologies, raising urgent demand for reliable evaluation towards health performance degradation of safety-critical assets. To meet the demand, remaining useful life (RUL), which aims at predicting the accurate failure time before a system failure happens [1]-[3], has garnered tremendous interest in various fields but not limited to rolling bearing [4], aircraft engine [5], high-speed train [6], etc. Consequently, there has been a substantial amount of work towards escalating RUL estimation accuracy by using in-depth process mechanisms \cite{Ref7}, \cite{Ref8}, pure machine learning models \cite{Ref9}, and hybrid modeling approaches \cite{Ref10}.

In its infancy of RUL estimation, the first principle models gain insightful degradation mechanism when mechanical systems are straightforward. Nowadays, the increasing mechanical complexity makes it challenging and impossible for domain experts to model mechanical dynamics accurately. With the rapid progress in sensor technology and advanced machine learning algorithms over the last decade, RUL estimation has witnessed the accelerating turns from physics-based methods towards data-driven intelligent models. Especially since 2016, there has been an upsurge in RUL estimation models employing deep neural networks (DNN). Although initially designed for computer vision and natural language processing, DNN and their sophisticated variants have been extensively used in machinery RUL estimation fields in light of their merits in feature representation and information inference.

Among DNN, convolutional neural network (CNN) and long-short term memory network (LSTM) are popular approaches for RUL estimation. The major merit of CNN lies in the ability upon the deep feature representation, such as exploiting the difference in the nose, eye, and mouth from portrait images for classification. In contrast, LSTM is good at capturing the long-term temporal correlation for sequential learning, making it naturally feasible for the RUL estimation task.  Taking raw time series from a multivariate engine system as inputs, Babu et al. \cite{Ref13} firstly attempted to establish a model using two convolutional layers for RUL estimation purpose. A deep CNN with more convolutional layers was proposed by Li et al. \cite{Ref14} to boost estimation accuracy further. Instead of using CNN, Zheng et al. \cite{Ref15} has created a new class of RUL estimation models based on LSTM, taking advantage of the sequential learning ability of LSTM. The methods mentioned above exhibit the common use of raw measurements during offline training and online application phases. That is, they are end-to-end learning models without leveraging feature engineering. Although end-to-end learning achieves enormous success in scenarios with massive data, such as natural language processing, image classification, etc., the capability may be weakened for industrial systems, where the amount of run-to-failure time series is limited. To address this concern, embedding feature engineering into advanced DNN, referred to as hybrid models, provides a bridge to convey statistical knowledge and enhance interpretability. Cheng et al. \cite{Ref16} constructed degradation indicators from raw data through empirical mode decomposition and then employed CNN to learn the hidden degradation patterns for final estimation. Wu et al. \cite{Ref17} used differential features of the raw data into the LSTM based-RUL estimation model. Recently, Chen et al. \cite{Ref18} introduced an attention mechanism into LSTM to assign more weights for crucial samples and features, trying to find the features most related to degradation procedure.

To date, capsule network (CapsNet) proposed by Sabour et al. \cite{Ref11} has been introduced to carry out the equivariant representation of entities in images. Minor changes in spatial relationships of local entities will be recognized during the training procedure, rather than ignorance in CNN. This capability is gained through the novel concept of capsules, each of which is a locally invariant group of neurons to recognize the presence of entities and encode their properties into a vector output. CapsNet outperforms CNN and LSTM on experimental data from the aircraft engine system \cite{Ref12}, attributed mainly to the high-dimensional feature representation. This further motivates us to explore the benefits of CapsNet for RUL estimation; however, some overlooked challenges in current approaches need to be clarified.

Machinery processes hold complex dynamics in normal status, mainly including temporal dynamics and slow-varying dynamics. From the viewpoint of causal analysis, the slow nature of raw materials fluctuations, operating condition changes, and minor disturbances in mechanical systems serve as the fundamental causes responsible for the slow-varying nature observed in measurements. The slow-varying dynamics exposes the inner strength of mechanical systems, corresponding to low-frequency components located in the power spectral density. When inherent sources with significant auto-correlations drive the measurable signals, slow-varying dynamics could reveal mutually uncorrelated components with varying frequencies. Moreover, degradation behaviour of mechanical systems always follows a long-term and gradual procedure, leading to an immediate and close impact on the slow-varying space derived from the normal status. As such, the slow-varying features will deviate from their normal patterns due to the adverse consequences of faults, raising the following concerns that have not been discussed yet:

\begin{itemize}
\item Current feature engineering ways for RUL estimation are trapped in steady variations from measurements, such as the mean, trend trajectory, principal components, etc. As such, these features fail to convey in-depth information that drives the system dynamically. Specifically, the importance of slow-varying dynamics, which refers to the variance of concerned features changes slowly in normal patterns, has been rarely unveiled for modeling.
\item Although the importance of temporal correlation in sequential data is widely recognized, CapsNet is incapable of capturing long-term temporal dynamics due to its inherent network structure. This may cap the full ability of CapsNet, as it is not designed initially for RUL estimation purposes. Out of this consideration, the redesign of CapsNet by taking long-term temporal capability into account may enhance RUL prediction performance.
\item As an emerging network, the study of CapsNet is still at its infant phase, and how to speed up the network configuration and save hyper-parameters tuning time deserves more effort. Lacking clear guidance for network configuration hinders pursuing an efficient CapsNet-based RUL model and consumes extensive time for hyper-parameters tuning.
\end{itemize}

Aiming to boost the estimation accuracy and simplify the network configuration, we propose a Slow-varying Dynamics assisted Temporal CapsNet (SD-TemCapsNet) for RUL estimation, consisting of three sequential parts. First, slow-varying dynamics concerned feature engineering is designed for the first time. The extracted slow features are mixed with raw measurements to construct hybrid features. To this end, hybrid features could be processed into data frames as inputs through data slicing. Second, a newly designed SD-TemCapsNet captures both slow-varying dynamics and long-term temporal dynamics of run-to-failure time-series in an iterative learning way. The two successive capsule layers hierarchically capture the local correlation and spatial relationship of data frames, transmitting low-level local features into high-level global features. Moreover, the introduction of LSTM accommodates the long-term temporal correlation between data frames. Third, an in-depth analysis of slow-varying dynamics accelerates the tuning of crucial hyper-parameters of SD-TemCapsNet. Through experiments on two representative machinery processes, SD-TemCapsNet offers remarkable results compared to current methods, illustrating the incremental benefits by considering slow-varying dynamics. The main contributions of this work are summarized as follows:

\begin{itemize}
\item We take slow-varying dynamics of run-to-failure time-series into consideration for machinery RUL estimation, contributing to an innovative feature engineering to boost estimation accuracy and gain insightful understandings;

\item We devise a novel SD-TemCapsNet for RUL estimation purpose, possessing the temporal learning ability via redesigned network structure and training procedure;

\item We put forward an operational solution to speed up the network configuration of SD-TemCapsNet, saving much tuning time and avoiding tedious network adjustment regarding hyper-parameters;

\item We verify the potency of the proposed method through extensive experiments, demonstrating the escalated estimation accuracy with both non-cyclic degradation and cyclic degradation processes.
\end{itemize}

\begin{figure}[!ht]
\centering
\includegraphics[scale=0.5]{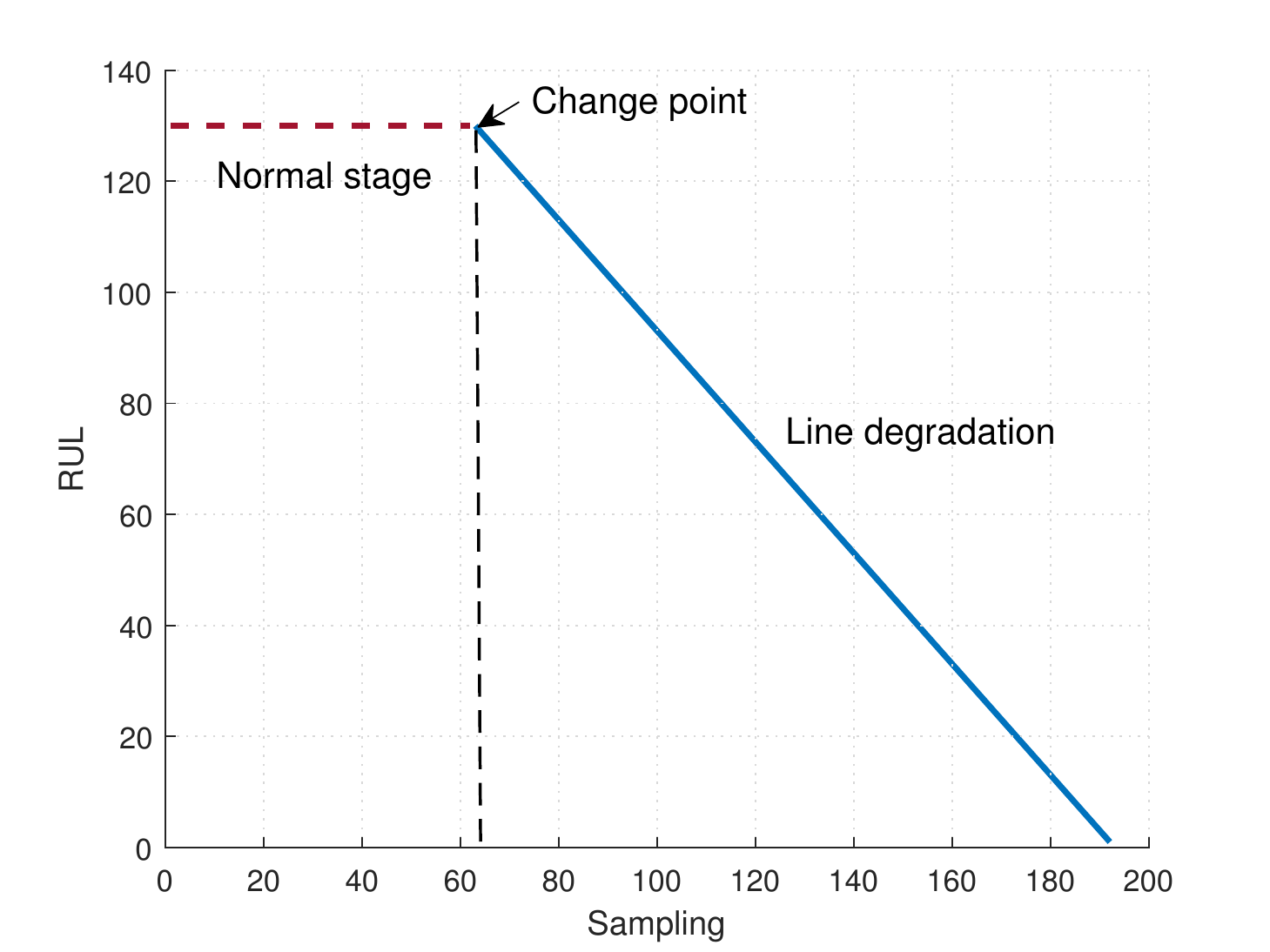}\\
\caption{Illustration of the piece-wise health index for aircraft engine system.}
\label{FIG1}
\end{figure}

\begin{figure*}[!htb]
\centering
\includegraphics[scale=0.39]{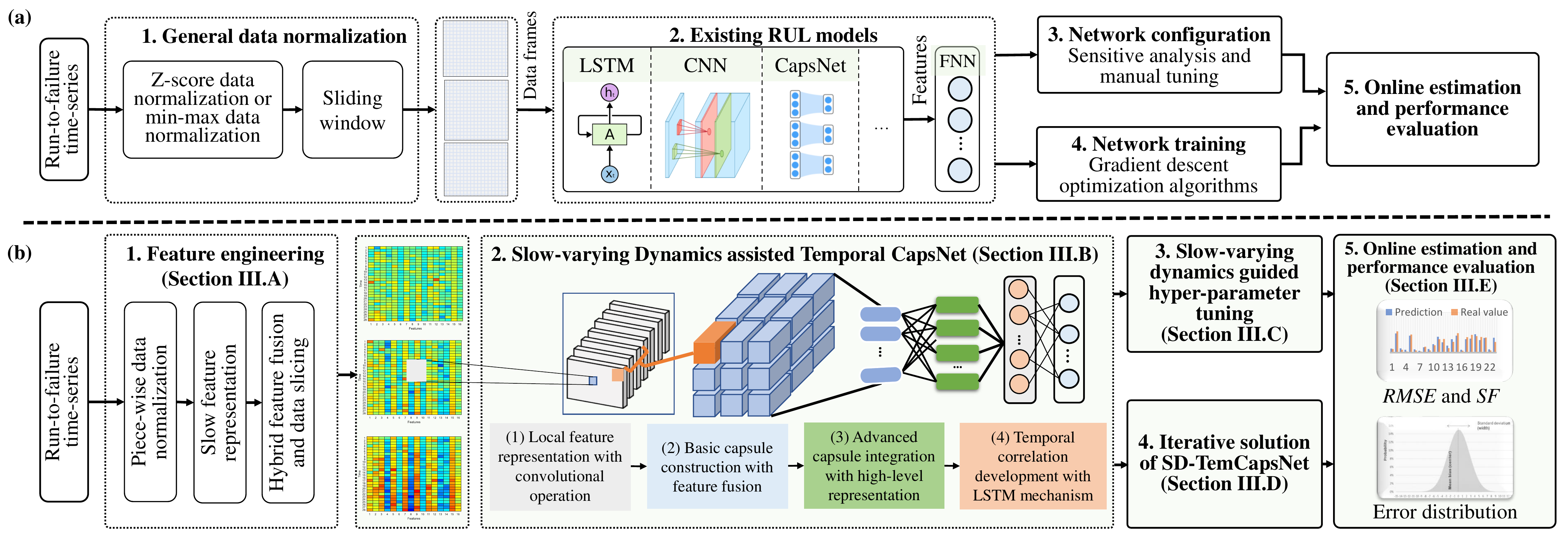}
\caption{Summary and comparison of system architectures between (a) the existing RUL estimation models and (b) the proposed SD-TemCapsNet model.}
\label{FIG2}
\end{figure*}

The structure of the remaining parts is organized as follows: Section II describes the data structure used for RUL estimation. Section III details the proposed SD-TemCapsNet with feature engineering, model architecture, and hyper-parameter tuning. Section IV illustrates systemic comparisons between the proposed method and the existing approaches. Section V concludes this work and provides future research.

\section{Data Description with Piece-wise Function}
Assuming a multivariate machinery process has $J$ variables, a complete run-to-failure time-series is denoted as data matrix $\mathbf X_c$ of dimension $J \times K_c$, where $K_c$ is the sequence length for the $c^{th}$ component throughout its service time. Usually, a machinery system will operate normally and then step into the degradation stage after a period of usage. Typically, a piece-wise linear RUL target function \cite{Ref29} has been adopted to act as the health index and separates the normal stage from the degradation stage below,

\begin{equation}
HI(k) = \left \{
\begin{aligned}
& RUL_{max},  & {k \leqslant K_{cp}} \\
& RUL_{max} - (k - K_{cp}),  & {k > K_{cp}}
\end{aligned}
\right.
\label{Eq1}
\end{equation}
where symbol $k$ is the time index, $K_{cp}$ is the change point between the normal stage and the degradation stage, and $RUL_{max}$ is the maximum RUL to avoid over-estimation.

Fig. \ref{FIG1} visualizes the trend of the health index for a non-cyclic degradation process \cite{Ref10}. Correspondingly, the ground truth of the health index is marked as $\mathbf y_c(1 \times K_c)$, which keeps constant in the normal stage and then linearly decreases after stepping into the degradation stage. It is worth noting that $K_{cp}$ is crucial to construct the health index. If expert knowledge is available to find $RUL_{max}$ \cite{Ref12}, $K_{cp}$ could be deduced inversely. Alternatively, multivariate statistical process analysis is capable of identifying the specific value of $K_{cp}$ by checking the process variations \cite{Ref30}, temporal dynamics \cite{Ref31}, etc.

\section{RUL Estimation using Slow-varying Dynamics Assisted Temporal CapsNet}
The basic structures of existing methods [12]-[18] are summarized in Fig. \ref{FIG2}(a) for straightforward comparison. The improvements of the proposed method SD-TemCapsNet are comprehensive, as shown in Fig. \ref{FIG2}(b). It includes feature engineering, details of the proposed SD-TemCapsNet, followed by the hyper-parameters tuning and iterative solutions.

\subsection{Feature Engineering with Slow-varying Dynamics}
\subsubsection{Piece-wise Data Normalization} We design a piece-wise data normalization by separately considering the normal status $\mathbf X_{n, c}$ and the degradation status $\mathbf X_{d, c}$ in the $c^{th}$ component, where the subscript $n$ indicates the normal stage and the subscript $d$ presents degradation stage. Assuming a number of $C$ failure components are available, a series of matrixes $\mathbf X_{n, 1}$, $\mathbf X_{n, 2}$, $\cdots$, $\mathbf X_{n, C}$ and matrixes $\mathbf X_{d, 1}$, $\mathbf X_{d, 2}$, $\cdots$, $\mathbf X_{d, C}$ will be arranged, corresponding to normal and degradation stage of each component. An integrated data matrix $\mathbf X_n$ is obtained through stacking matrixes $\mathbf X_{n, 1}$, $\mathbf X_{n, 2}$, $\cdots,$ $\mathbf X_{n, C}$ along the variable-wise direction, where the dimension of variables kept unchanged. Performing z-score normalization on $\mathbf X_n$, its mean and standard variation are calculated as $\mathbf u$ and $\mathbf \Lambda$ with $J$ dimensions. Applying z-score normalization for each degradation data matrix, the normalized data could be calculated as,
\begin{equation}
\begin{aligned}
\widetilde{\mathbf {X}}_{d,c}(k)= \frac {\mathbf X_{d,c}(k) - \mathbf u} {\mathbf \Lambda}
\end{aligned}
\label{Eq2}
\end{equation}
where $\mathbf X_{d,c}(k)$ is the $k^{th}$ sample of matrix $\mathbf X_{d,c}$. For brevity, the normalized matrix is still denoted as ${\mathbf {X}}_c$.

\textit {Remark:} Min-max normalization and z-score normalization are typical ways to handle the negative influence of variable scale for run-to-failure time series. However, few studies have pointed out the disadvantages of using normalization information with the complete run-to-failure time series. The min-max normalization unexpectedly reduces the dissimilarity between normal and faulty data at the early degradation stage. For z-score normalization, the mean locates at the middle of the degradation stage, and the variation is relatively large. That is, the time-series distributed at the normal status vary in a stationary behaviour. In contrast, run-to-failure time series follow a non-stationary trend caused by faults. The benefits of piece-wise normalization are twofold. First, minor changes at the incipient degradation stage could be amplified compared to min-max normalization. Second, it is likely to enlarge the degradation signs of measurements at the beginning and ending time in comparison with z-score normalization.

\subsubsection{Slow Feature Representation}
Our proposed method gives particular attention to the slow-varying characteristic of time-series in the normal operation stage. Furthermore, this dynamics may be significantly influenced in the degradation procedures along with the changes of steady variation.

The slow features are conceptually defined as latent variables whose varying speed of steady variations change slowly \cite{Ref19}-[24]. And they would convey primary information embodied in time series. As such, a series of slow features are optimized by minimizing the temporal variation as follows,

\begin{equation}
\begin{array}{l}
{\Delta(\mathbf{s}_{i}):=\langle\mathbf{\dot s}_{i}^{2}\rangle} \\ [1mm]
{{s.t.} \langle\mathbf{s}_{i}\rangle= 0} ; {\langle\mathbf{s}_{i}^{2} \rangle= 1}; {\langle\mathbf{s}_{i} \mathbf{s}_{j}\rangle= 0 \quad \forall j<i}\end{array}
\label{Eq3}
\end{equation}
where $\dot {\mathbf s}_i$ is the first-order temporal difference of the $i^{th}$ slow feature $\mathbf {s}_i$. The first two constraints in Eq. (3) ensure the mean and variance of $\mathbf s_i$ are zero and unit, respectively. Besides, the third constraint removes redundant information among slow features by ensuring the independence between each other.

To pursue clear interpretability, a series of linear transformations $\bm \omega$ are preferred to construct the slow-varying space from the normal data $\mathbf X_n$. Correspondingly, slow features in normal data are defined as $\mathbf s=\mathbf {X}_n \bm \omega$. The collection of weighting vectors $\bm \omega$ can be calculated by solving the eigenvalue decomposition problem as follows,
\begin{equation}
{\bm \Sigma}_{\delta} {\bm \omega}={\lambda} {\bm \Sigma_n} {\bm \omega}
\label{Eq4}
\end{equation}
where ${\bm \Sigma_n}$ is the covariance of $\mathbf X_n$; ${\bm \Sigma}_{\delta}$ is the covariance of $\dot {\mathbf X}_n$, in which $\dot {\mathbf X}_n$ is the first-order temporal difference of $\mathbf X_n$; $\lambda$ is the eigenvalue reflecting the slowness of slow features, which is sorted in descending order. In supplement, the detailed procedure paves the path for the eigenvalue decomposition.

Slow features are evaluated and selected according to the criterion of slowness. Assuming the first $P$ latent variables in Eq. (\ref{Eq4}) are retained, the corresponding weighting matrix is denoted as $\mathbf \Omega_s = [\bm \omega_1, \bm \omega_2,...,\bm \omega_P]$, constructing the slow-varying subspace. The remaining vectors $[\bm \omega_{P+1}, \bm \omega_{P+2},...,\bm \omega_J]$ form the residual space $\mathbf \Omega_r$.

Projecting degradation part $\mathbf X_{d,c}$ of $\mathbf X_c$ on the slow-varying subspace $\mathbf \Omega_s$, the derived slow features in degradation stage of the $c^{th}$ component are expressed as follows,

\begin{equation}
{\mathbf S}_{d,c}={\mathbf X}_{d,c} {\mathbf \Omega}_s
\label{Eq5}
\end{equation}

It is worth noting that the slow-varying space $\bm \Omega_s$ is decomposed only using the normal data $\mathbf X_n$. Then $\bm \Omega_s$ will be directly used in the degradation stage, yielding the slow features sensitive to the fault evolution.

\subsubsection{Hybrid Feature Fusion and Data Slicing} 
\begin{figure}[H]
\centering
\includegraphics[scale=0.4]{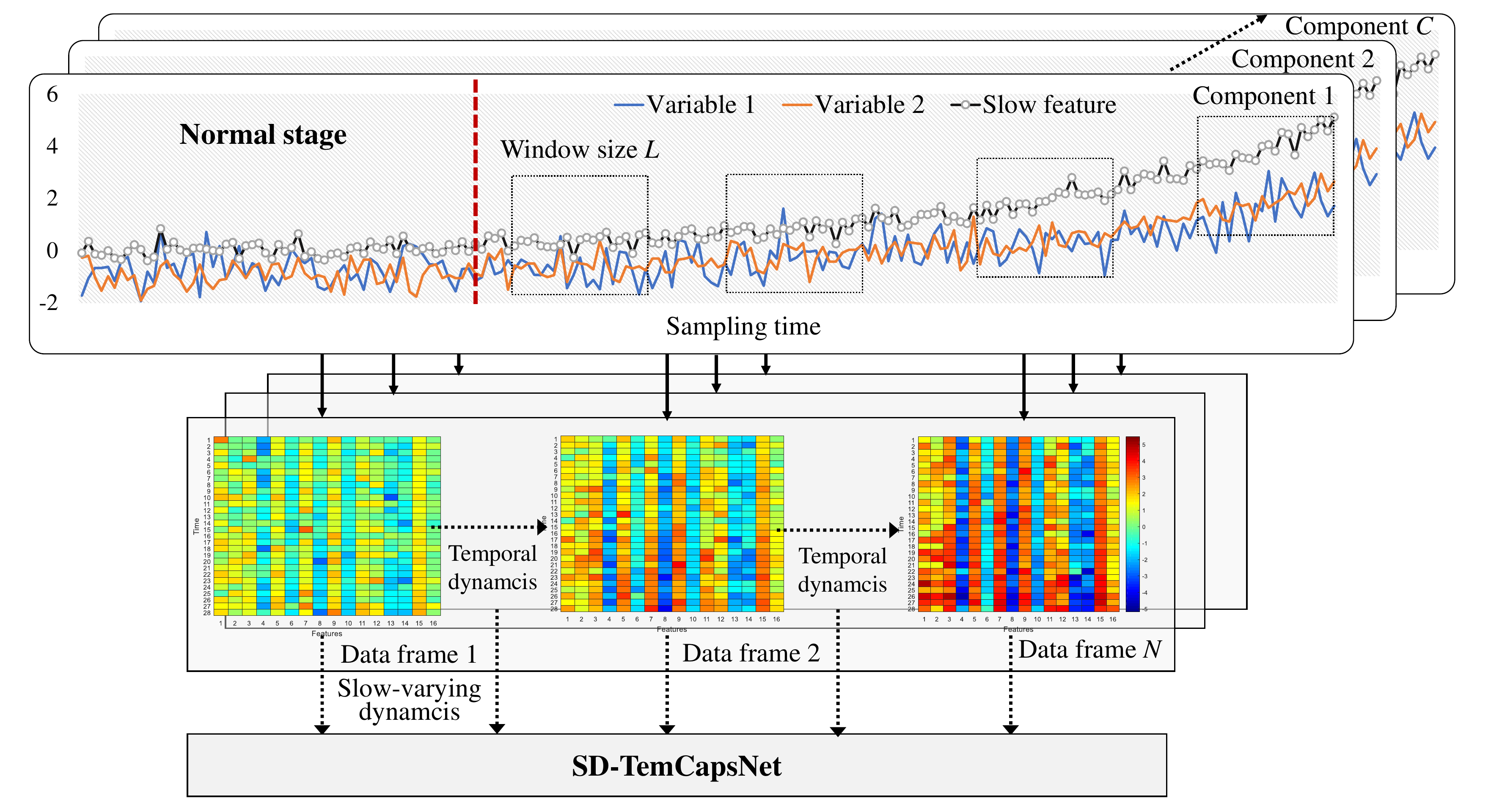}
\caption{The illustration of data slicing on run-to-failure time-series with temporal dynamics and slow-varying dynamics.}
\label{FIG3}
\end{figure}

Hybrid features of each run-to-failure are constructed by staking $\mathbf X_{d,c}$ and $\mathbf S_{d,c}$ together along the variable direction, where the number of samples keeps the same. For clarity, the hybrid features are denoted as $\mathbf T_c = [\mathbf X_{d,c}, \mathbf S_{d,c}] (c\in [1, C])$. To increase the number of samples for network training, data slicing is conducted on $\mathbf T_c$ to divide a long run-to-failure time-series into a series of data segments with dimensions $L \times (J+P)$. Fig. \ref{FIG3} illustrates the detailed data slicing procedure and further visualizes the integration of temporal dynamics and slow-varying dynamics. It should be noticed that the window length $L$ relates to the temporal correlation between the past data and future data. Thus, the specific values of $L$ can be calculated according to the sample autocorrelation function of slow features from the degradation data using two standard deviations \cite{Ref22}. Consequently, the produced matrices are image-like and referred to as data frames in the following sections, serving as the basic data unit.

\subsection{Slow-varying Dynamics Assisted Temporal CapsNet}
The $i^{th}$ data frame $\mathbf T_{i,c}$ generated from $\mathbf T_c$ are input data to predict the real RUL value, i.e., label, yielding estimations as output. Apart from the input and output layers, the proposed SD-TemCapsNet hierarchically consists of four parts, which are local feature representation, basic capsule construction, advanced capsule integration, and temporal correlation development. The details of each part and its associated computation processes are illustrated as follows.

\subsubsection{Local Feature Representation with Convolutional Operation} The typical CNN layer is adopted to extract local features through convolution kernels. Weights of each kernel are randomly given and kept the same when kernels slide on input data frames. The generated two-dimensional data matrix is usually recorded as a feature map corresponding to each kernel. Multiple feature maps will be generated with different kernels. For the $m^{th}$ kernel ${\mathbf \Phi}_m$, the associated feature map $\mathbf F_m$ could be derived as,
\begin{equation}
\mathbf F_m  = tanh(\Phi_m \ast \mathbf T_{i,c} + \mathbf b_m)  \quad m \in [1, M] \\
\label{Eq6}
\end{equation}
where $\mathbf b_m$ is the $m^{th}$ bias term, $\ast$ is the convolutional operation, $M$ is the number of kernels, and $tanh(\cdot)$ is the hyperbolic tangent function.

\subsubsection{Basic Capsule Construction with Feature Fusion} This layer aims to transform the scalar-based outputs into vector-based features in the form of capsules. Technically, a capsule is a collection of adjacent but non-repetitive convolutional units. From the data analysis perspective, a capsule is the smallest element of the whole feature space, and it could be combined with each other to construct high-level features with higher dimension and presentation capability.

Feature fusion is conducted along filter direction to group values at the same position. Feature maps in the last step are grouped into $M/D$ matrixes in a three-dimension way, where $D$ is the dimension of basic capsules. Flattening the three-dimension matrix by keeping the dimension $D$ unchanged, a series of basic capsules $\mathbf u_i$ will be generated. In this way, the ability to extract local features used to describe a data frame will be greatly increased and freed from the limitation of the number of filters. For instance, for an image with dimensions $28 \times 28$, there will be 64 feature maps if the number of filters is 64. Assuming that the value of $D$ is 4 and the size of filters is (1, 2) with stride (1, 2), we could have 6276 basic capsules.

\subsubsection{Advanced Capsule Integration with High-level Representation} With the generated basic capsules in the last layer, the advanced capsule layer is developed with dual considerations. First, although basic capsules allow describing features on a small dimension scale, some may be redundant, and removing unimportant ones helps avoid disturbing the performance. Second, advanced capsules with high-dimension will be introduced to describe high-level features, which are the combinations of shallow features given by basic capsules. From the above analysis, the dimension of advanced capsules will be much higher than that of the basic capsules, which is usually given as 2$D$. Naturally, the number of advanced capsules will be largely reduced.

In practice, we use a fully connected layer with definite neurons to construct the advanced capsules. The adjacent neurons will be assigned into the same group to represent the $j^{th}$ advanced capsule $\mathbf v_j (1 \times 2D)$.

\subsubsection{Temporal Correlation Development with LSTM} 
\begin{figure}[H]
\centering
\includegraphics[scale=0.5]{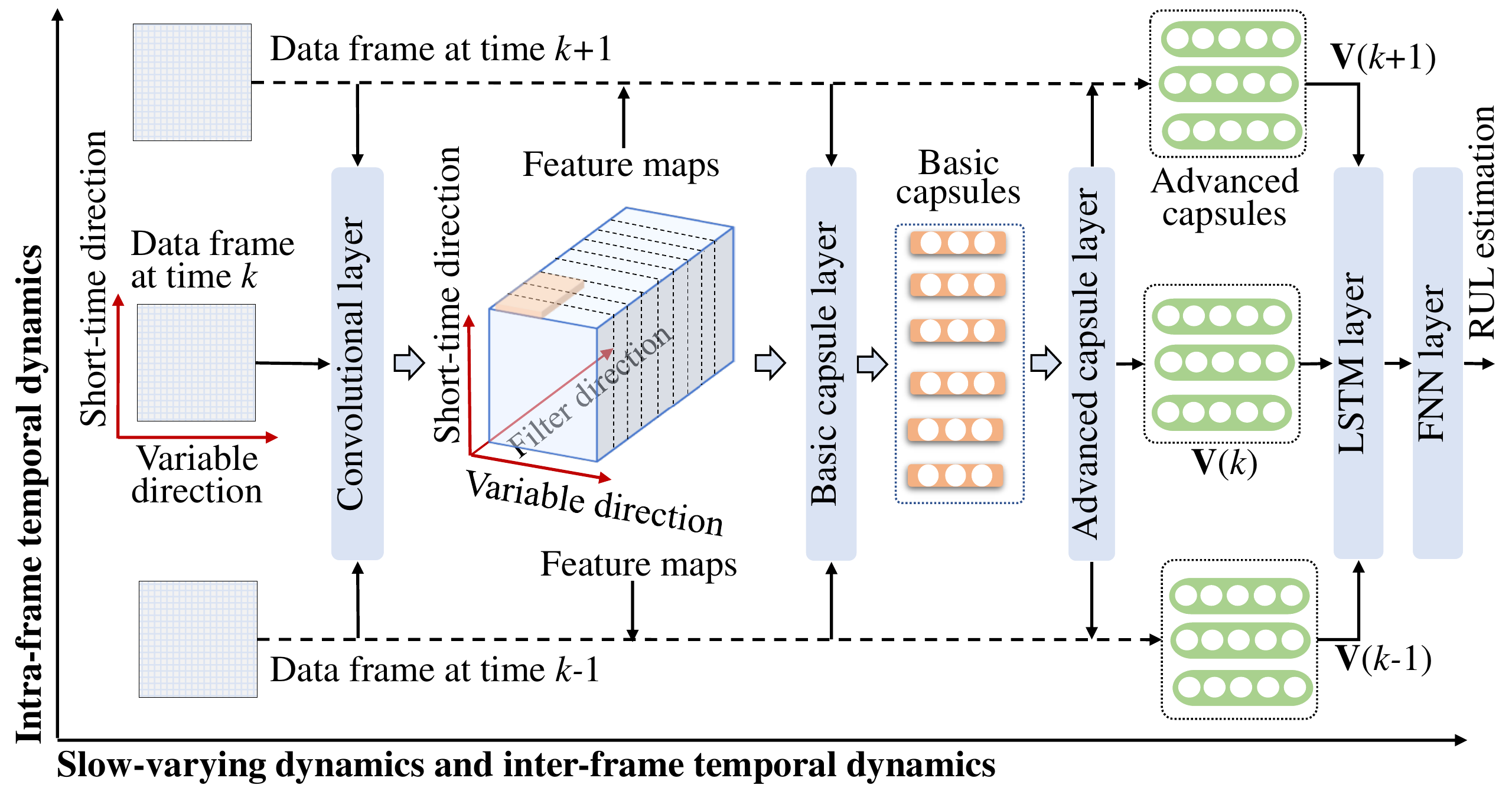}
\caption{The data flow and network structure of the proposed SD-TemCapsNet.}
\label{Fig4}
\end{figure}

The internal temporal and spatial correlation of a data frame can be well captured through the given multi-layer feature representation. To consider the temporal correlation of failure degradation among data frames, LSTM has been designed for sequential learning by introducing a memory mechanism into the neural network. As shown in Fig. \ref{Fig4}, passing the derived features $\mathbf V(k)$, $\mathbf V(k+1)$, $\cdots$, $\mathbf V(k+K)$ over LSTM layers connects the inherent correlations between data frames.

Multiple linear regression layers with a proper dropout probability follow the LSTM layers to regress the highly abstract feature with labeled data.

\subsection{Slow-varying Dynamics Guided Hyper-parameters Tuning}
The network configuration has always been a challenge for DNN. Especially, as an emerging approach, limited studies are available to direct how to tune CapsNet. In light of the ability of slow features in explaining first-order systematical dynamics, they are employed to guide hyper-parameters of SD-TemCapsNet, as summarized in Table \ref{Table1}.

\begin{table}[H]
	\renewcommand{\arraystretch}{1.2}
	\centering
	\caption{Summary of the hyper-parameters in SD-TemCapsNet.}
	\scriptsize
	\begin{tabular}{p{0.8cm} p{3.3cm} p{3.5cm}}
			\hline
			\hline
			\textbf {Type} & \textbf {Parameter name} & \textbf {Recommended value} \\
			\hline
			\multirow{9}{*}{Definite} & \multirow{3}{*}{Window length $L$} & Finding the change point regarding slowness from the singular value $\lambda$ calculated from Eq. (4) \\
			\cline{2-3}
			& \multirow{3}{*}{No. of slow features {$P$}} & Sample autocorrelation function of slow features from the degradation data using two standard deviations\\
			\cline{2-3}
			& No. of advanced capsules & The value of $P$ \\
			\cline{2-3}
			& Dimension of advanced capsules & The value of $J$ \\
			\cline{2-3}
			& Dimension of basic capsules & The value of $D=\frac {1} {2} (P+J)$ \\
			\hline
			\multirow{3}{*}{Tunable} & No. of filters ($M$) & Sensitive analysis \\
			\cline{2-3}
			& No. of basic capsules & The value of $M/D$ \\
			\cline{2-3}
			& No. of LSTM cells & Sensitive analysis \\
			\hline
			\hline
	\end{tabular}
\label{Table1}
\end{table}

For multivariate statistical analysis modelling, the remarkable explainability of slow features for regressing performance indicators in linear cases is exploited \cite{Ref19}. The essence of slow features has been integrated into advanced capsules through data fusion in a non-linear way, sharing similarities in feature presentation. Assigning each element of the output vector a specific meaning, each dimension of advanced capsules is a non-linear combination of original measurements. Here, we could give the number of slow features to the number of advanced capsules, i.e., $P$, treating advanced capsules as the non-linearly slow features. Likewise, the dimension of advanced capsules has the same value as that of the number of variables, i.e., $J$. For simplicity, the dimension of basic capsules will be half of the advanced capsules, i.e., $D=1/2(P+J)$.

Besides, two independently tuneable parameters need to be determined using sensitivity analysis, including the number of filters $M$ and LSTM cells. The number of basic capsules depends on $M$, which can be determined as $M/D$. Although there is no clear theory to follow, we suggest that the number of filters and the number of LSTM cells are selected as the times of 8 according to our experience. Therefore, the sensitive analysis could start from the minimum of both parameters and stop when no improvement is observed.

\subsection{Iterative Solution of SD-TemCapsNet}
After giving specific values for hyper-parameters listed in Table \ref{Table1}, the proposed SD-TemCapsNet is trained in an iterative way, which integrates both the ``dynamic routing" mechanism \cite{Ref11} and gradient descend to update weights and bias of the network. Dynamic routing is mainly used for optimizing parameters of basic capsules and advanced capsules. Gradient descend aims at updating the weights and bias of LSTM.

With dynamic routing, the basic capsules that are highly related to the advanced capsules will be located. The specifics of dynamic routing are described below.

\subsubsection{Computation of Coupling Coefficients}
Assuming $b_{i, j}$ is the prior probability that a basic capsule $i$ coupled to an advanced capsule $j$, the normalized probability $c_{i, j}$ with consideration of the capsule $i$ to other advanced capsules is calculated as below,
\begin{equation}
c_{i, j}= \frac{\exp (b_{i, j})} {\sum_{k} \exp (b_{i, k})}  \quad \forall i, j
\label{Eq7}
\end{equation}
where $b_{i, k}$ is the prior probability that capsule $i$ coupled to advanced capsule $k$. It should be noted that the value of $b_{i, j}$ in the first iteration epoch is set to zero, in which each advanced capsule equally accepts probabilities of a basic capsule.

\subsubsection{Input of Advanced Capsules}

By weighting sum over all input vectors $\hat{\mathbf u}_{j \mid i}$ from the basic capsules, the input of an advanced capsule $j$ is calculated below,
\begin{equation}
\mathbf s_{j}=\sum_{i} c_{i, j} \hat{\mathbf u}_{j \mid i}
\label{Eq8}
\end{equation}
where $\hat{\mathbf u}_{j \mid i}=\mathbf W_{i, j} \mathbf u_{i}$, and $\mathbf W_{i, j}$ is the weight matrix to relate basic capsule $i$ and advanced capsule $j$.

\subsubsection{Calculation of Posterior Probability}
The initial coupling coefficients are refined by updating $b_{i, j}$ according to the following rule,
\begin{equation}
b_{i, j}=b_{i, j}+\hat{\mathbf u}_{j \mid i} \cdot \mathbf v_{j}
\label{Eq9}
\end{equation}
where $\mathbf v_j$ is the output of the $j^{th}$ advanced capsule, which is defined below,
\begin{equation}
\mathbf v_{j}=\frac{\|\mathbf s_{j}\|^{2}}{1+\| \mathbf s_{j}\|^{2}} \frac{\mathbf s_{j}}{\|\mathbf s_{j}\|}
\label{Eq10}
\end{equation}

Eq. (\ref{Eq10}) is named as the squashing function. On the basis of this, the short vectors shrink to approximately zero, and long vectors approach a length slightly below one. In this way, length of the output vector of a capsule stands for the probability of the existence of extracted local features.

The routing weights of basic and advanced capsules will converge after repeating Steps 1)-3) several times (e. g., 3). In comparison with CNN, the dynamic routing is easy to be optimized, and the computation complexity is reduced.

For final prediction purposes, the advanced capsules learned from a specific data frame are temporally connected through LSTM and passed into the associated FNN layers to predict the given ground truth as accurately as possible. For clarity, the details of LSTM are not shown here, but could be easily found in existing literature \cite{Ref18}. Using $F_{L}(\cdot)$ to stand for the learned relationships, RUL estimation at time $k$ could be estimated as below,
\begin{equation}
\hat {\mathbf Y}_k = F_{L}(\mathbf {V}_k)
\label{Eq11}
\end{equation}
where $\mathbf {V}_k$ is the collection of outputs from the advanced capsule layers for the $k^{th}$ data frame.

The loss function is defined as the sum of errors between the estimations $\hat {\mathbf Y}_k$ and the corresponding ground truth ${\mathbf Y}_k$ to guide the parameter adjustment of LSTM and FNN layers,
\begin{equation}
\min \limits_{F_{L}} E = \sum_{k=1}^{K} (F_{L}(\mathbf {V}_k) - {\mathbf Y}_k)^2
\label{Eq12}
\end{equation}

Algorithm 1 summarizes the detailed steps to train SD-TemCapsNet iteratively, consisting of two sequential parts. The CapsNet part is updated using the routing agreement, and the part of LSTM is updated with gradient descend. Adam optimization \cite{Ref23} is recommended to tune the weights and biases during training. To avoid overfitting, early stopping is applied when no performance improvements are observed in the training procedure.

\begin{algorithm}[H]
\caption{Iterative training of SD-TemCapsNet}
\scriptsize
\LinesNumbered
\KwIn{Data frame ${\mathbf T_{i,c}}$, corresponding ground truth $\mathbf Y_{c,i}$, and specific values of hyper-parameters listed in Table I}
\KwOut{Coefficients of CapsNet part $\mathbf \Theta_C$ and LSTM part $\mathbf \Theta_L$}
Randomly initialize $\mathbf \Theta_C$ and $\mathbf \Theta_L$ \\
\Repeat
{The given epochs are reached or early stopping when no improvements are observed on the validation dataset.}
{\Repeat (//Update the CapsNet part)
{Given iterations (here is 3)}
{Initialize $b_{i, j}$ (0 for the first iteration in the first epoch) and calculate $c_{i,j}$ according to Eq. (7); \\
Calculate the inputs of advanced capsules $\mathbf s_j$ according to Eq. (8); \\
Update $b_{i, j}$ according to Eqs. (9) and (10); \\
Generate outputs of advanced capsules $\mathbf v_j$; \\}
{Output the advanced capsules to the following LSTM layer;}\\
\Repeat (//Update the LSTM network)
{}
{Update parameter of LSTM $\mathbf \Theta_L$ to minimize, \\
$ E = \sum_{t=1}^{T} (F_{LSTM}(\mathbf {V}_t) - {\mathbf Y}_t)^2 $ \\
$\mathbf \Theta_L \leftarrow \mathbf \Theta_L + \nabla E(\mathbf \Theta_L)$ \\}
}
\end{algorithm}

\subsection{Online RUL Estimation and Performance Evaluation}
We perform piece-wise normalization as the first step for the incoming data frame $\textbf x_{new}(J \times L)$ and then obtain the final features by projecting the normalized data on the slow-varying space. Next, RUL estimation $RUL_{new}$ is calculated through feeding processed features into the well-trained SD-TemCapsNet. Two kinds of indices are employed to evaluate the performance of the proposed method comprehensively. The first one assesses the overall performance by treating all testing parts as a whole. The second one depicts the distribution of estimation errors accordingly.

\subsubsection{Overall Performance Indices} For a fair comparison, the commonly used indices root mean square error ($RMSE$) and scoring function ($SF$) measure the overall performance. $RMSE$ assigns the same weight for each part, and the average result is achieved. $SF$ gives a large punishment for the case when lagged estimation is observed \cite{Ref10}. Both values of $SF$ and $RMSE$ are smaller, the better.

\subsubsection{Error Distribution} Due to $RMSE$ and $SF$ are overall metrics by averaging all weighted estimation errors, a further comparison concerning the distribution of estimation errors is necessary. Specifically, the method with estimation errors distributed in the low-error bands will be preferred if two methods have a similar value of $RMSE$ or $SF$.

\section{Result and Discussion}
Two multivariate machinery systems are adopted for evaluating the performance of the proposed algorithm in this section. The first one is an aircraft engine \cite{Ref24}, \cite{Ref25}, and the second system is a milling machine used for manufacturing parts in a period manner \cite{Ref26}. Although both are machinery processes, the difference between these two systems can be observed from the measurements. A sample of the milling machine system is a two-dimensional data matrix rather than a vector collected from the aircraft engine system. 

\subsection{Experiments on Aircraft Engine System}
National Aeronautics and Space Administration (NASA) has launched a commercial modular aviation propulsion system simulation to simulate aircraft engine degradation procedure \cite{Ref24}. This system sequentially consists of four components: fan, compressor, combustor, and turbine, as shown in Fig. 5. In the system, closed control loops are deployed to adjust altitude, mach number, and throttle resolver angle during aviation. As a result, several operating conditions are simulated to manipulate system statuses, which are sensed by 21 sensors deployed at specific locations as shown in Fig. 5.

\begin{figure}[H]
\centering
\includegraphics[scale=0.5]{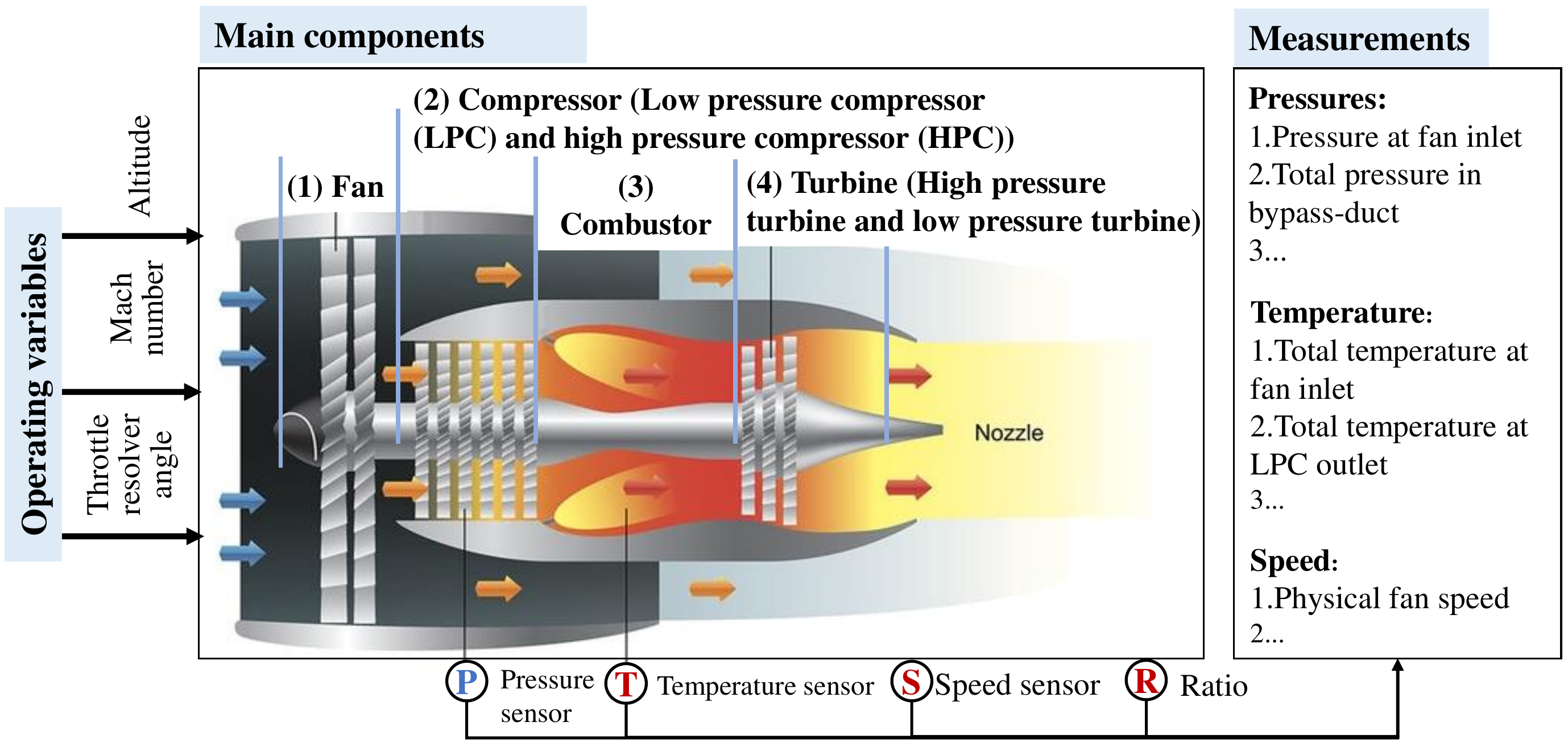} \\
\begin{center}
\caption{The system structure of the aircraft engine and measured variables.}
\end{center}
\label{Fig5}
\end{figure}

Table \ref{Table2} summarizes the primary information of four sets of experiment settings corresponding to various working conditions and faulty components. Corresponding datasets are collected as FD001, FD002, FD003, and FD004, respectively. FD001 and FD003 are generated with a single operation, and each of them contains 100 training trajectories and 100 testing trajectories. FD002 and FD004 experience more complex situations with random adjustment of controlled variables. The various working conditions and multiple faults lead to the different data distribution of each dataset. The ground truth of each engine trajectory is available for supervised learning. More information about measurements can be found in \cite{Ref24}.

\begin{table}[H]
\renewcommand{\arraystretch}{2}
\tiny
\caption{Fundamental information of C-MAPSS dataset.}
\label{Table five}
\begin{center}
\begin{threeparttable}
\begin{tabular}{c c c c c c}
\hline
\hline
\textbf{Dataset Name} & {FD001} & {FD002} & {FD003} & {FD004} \\
\hline
\textbf{Number of training set} & 100 & 260 & 100 & 248 \\
\hline
\textbf{Number of testing set} & 100 & 259 & 100 & 249   \\
\hline
\textbf{Number of operating conditions} & 1 & 6 & 1 & 6 \\
\hline
\textbf{Fault components} & High pressure compressor & High pressure compressor & High pressure compressor and fan & High pressure compressor and fan \\
\hline
\textbf{Length distribution of training set} & $\raisebox{-.5\height}{\includegraphics[scale=0.15]{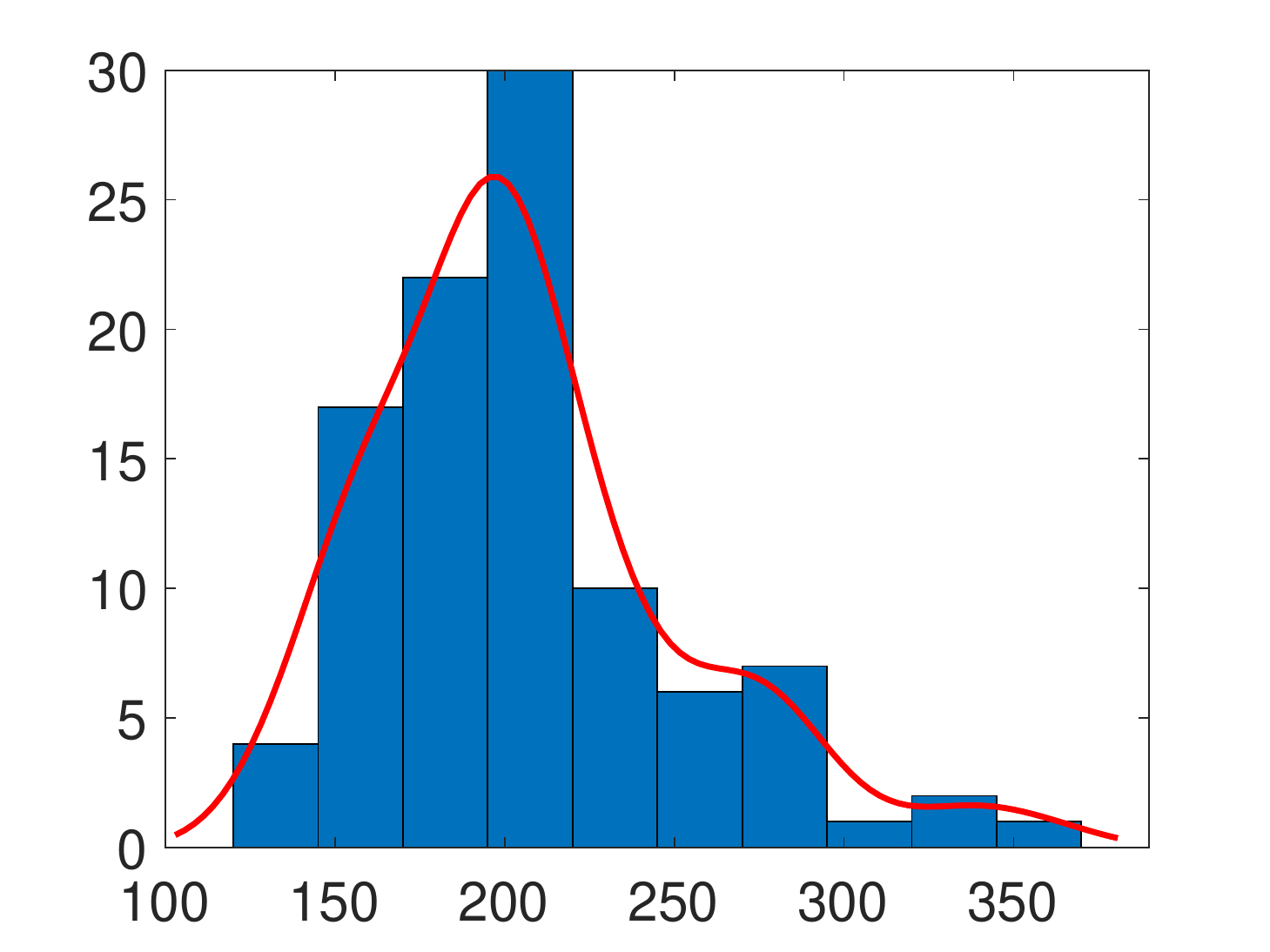}}$ & $\raisebox{-.5\height}{\includegraphics[scale=0.15]{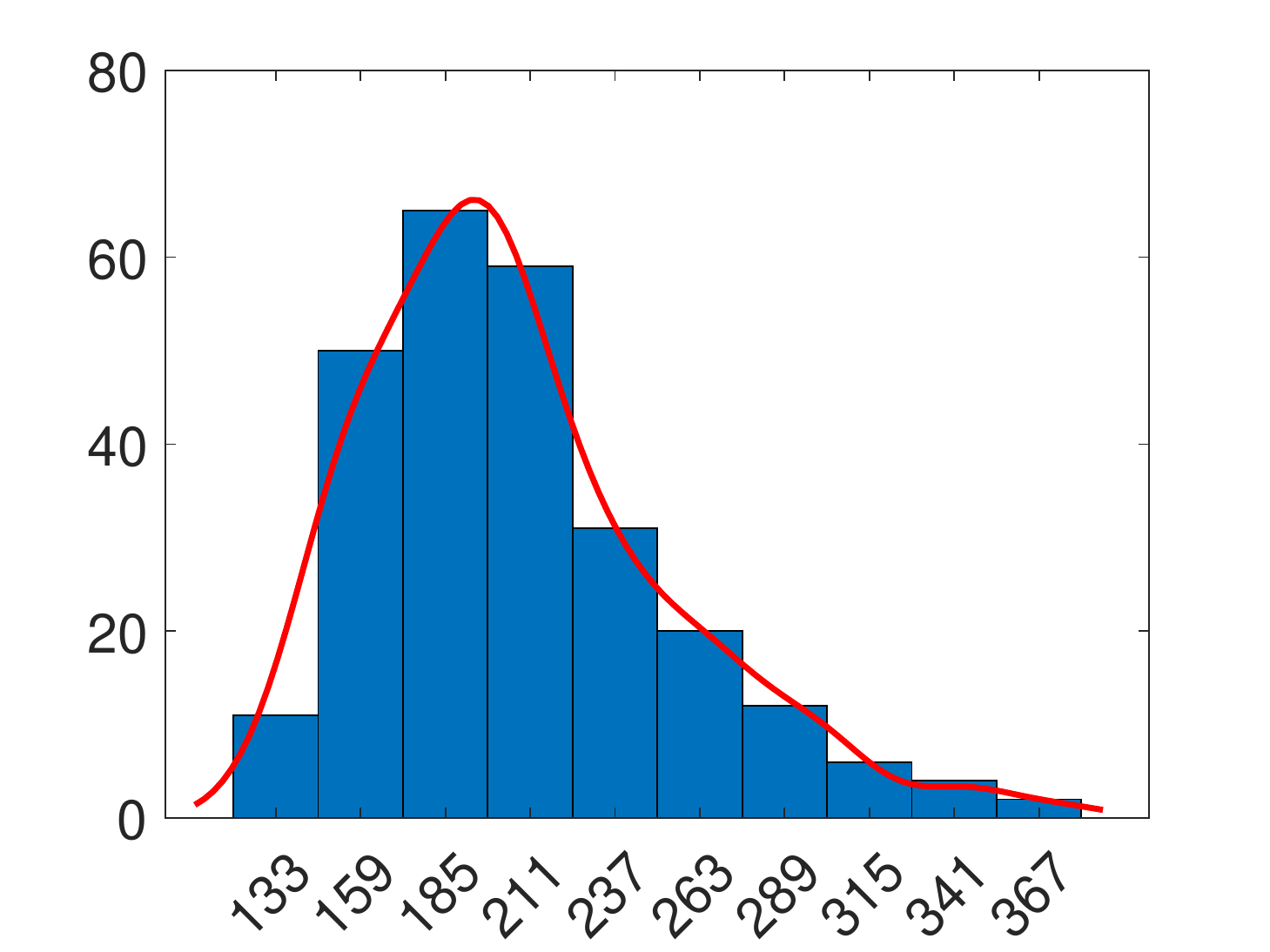}}$ & $\raisebox{-.5\height}{\includegraphics[scale=0.15]{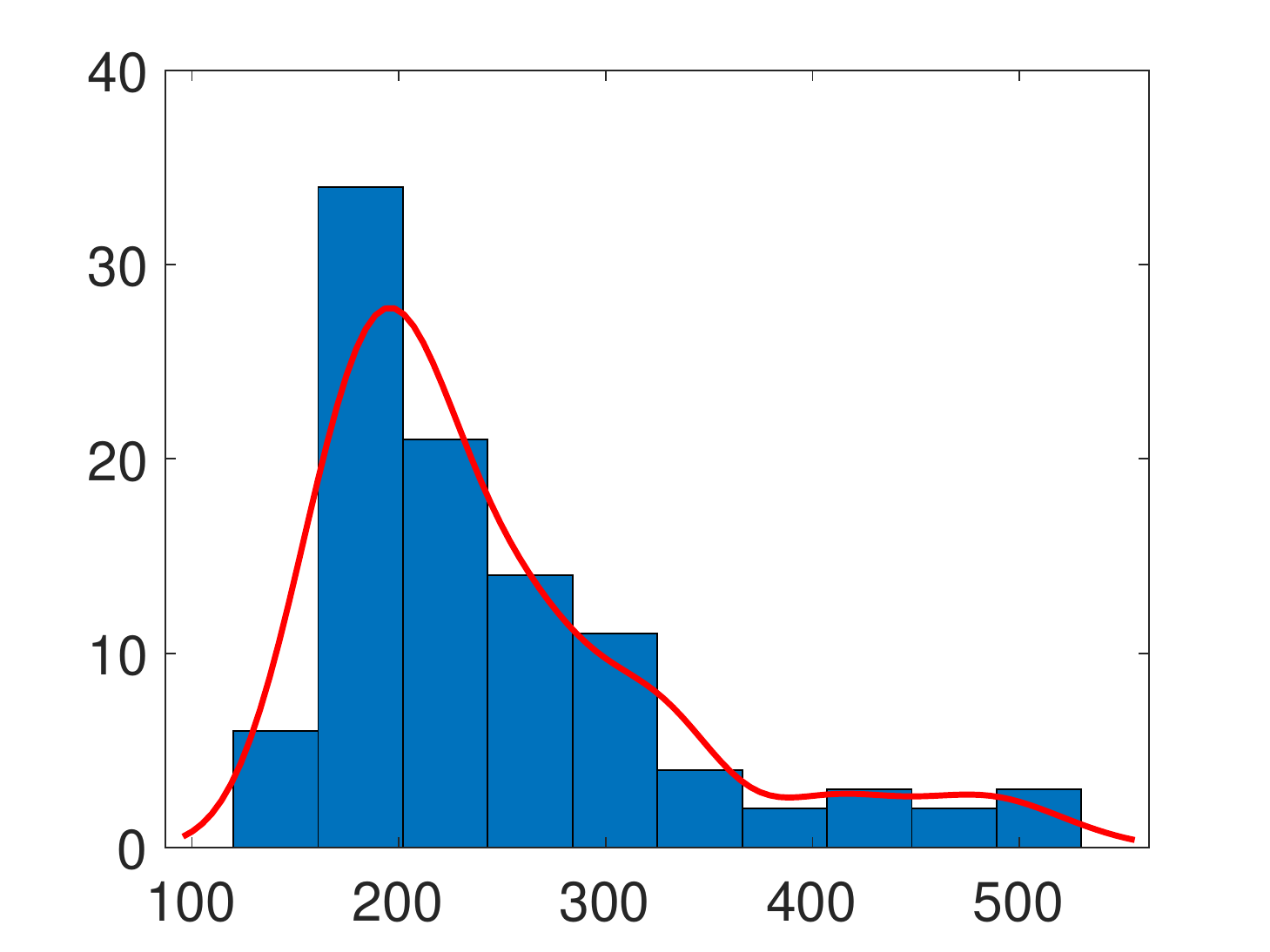}}$ & $\raisebox{-.5\height}{\includegraphics[scale=0.15]{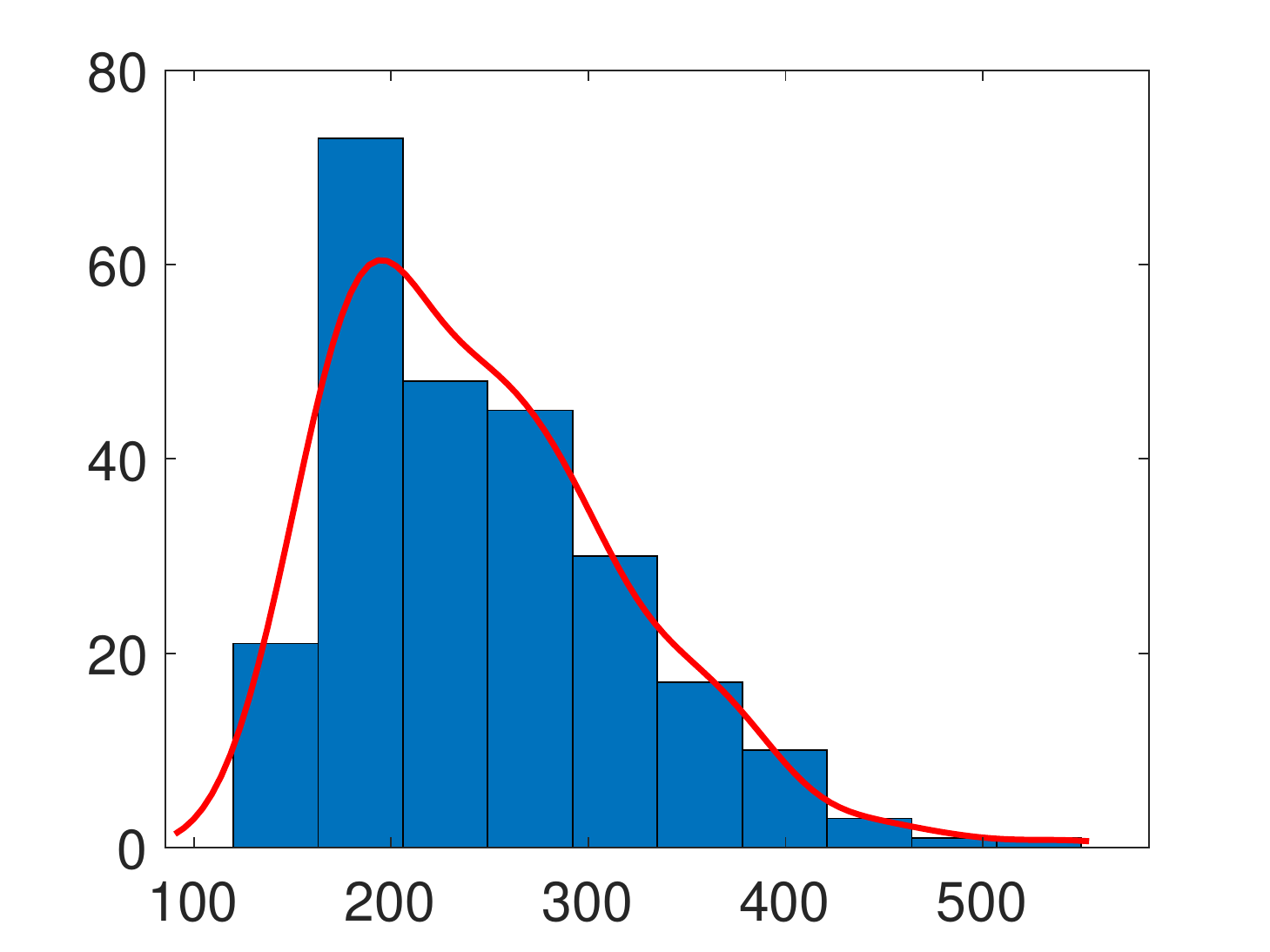}}$ \\
\hline
\textbf{Length distribution of testing set} & $\raisebox{-.5\height}{\includegraphics[scale=0.15]{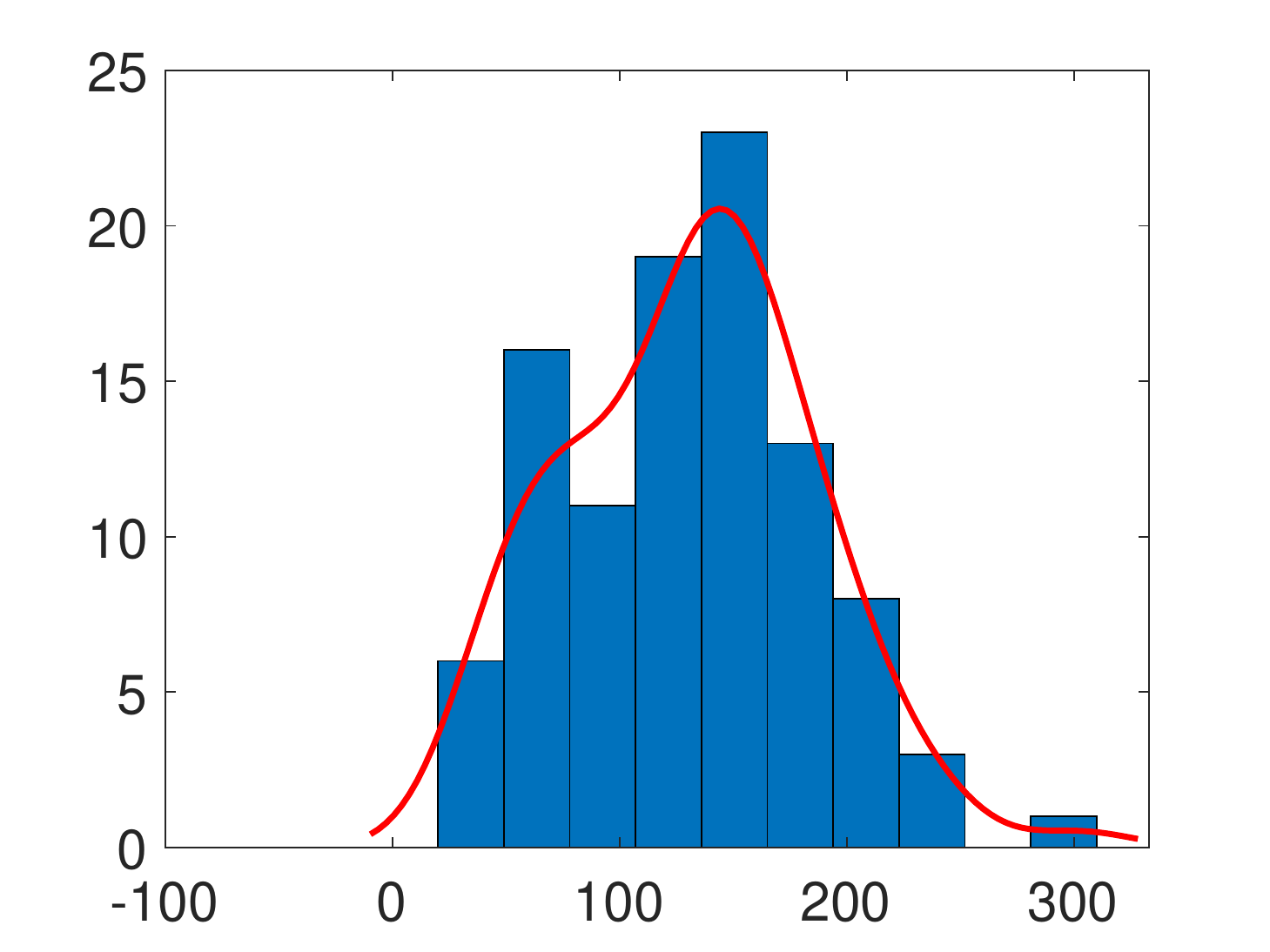}}$ & $\raisebox{-.5\height}{\includegraphics[scale=0.15]{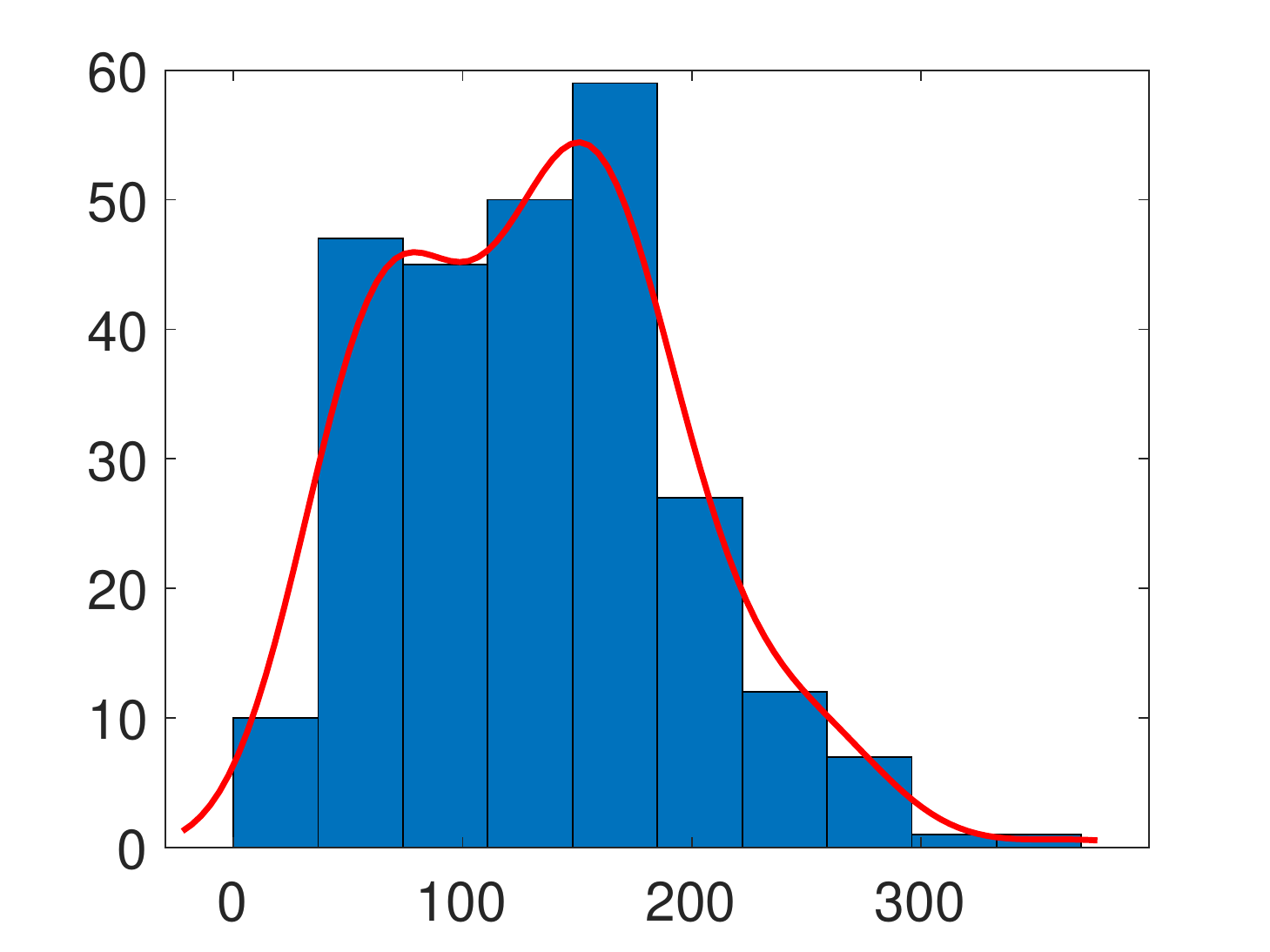}}$ & $\raisebox{-.5\height}{\includegraphics[scale=0.15]{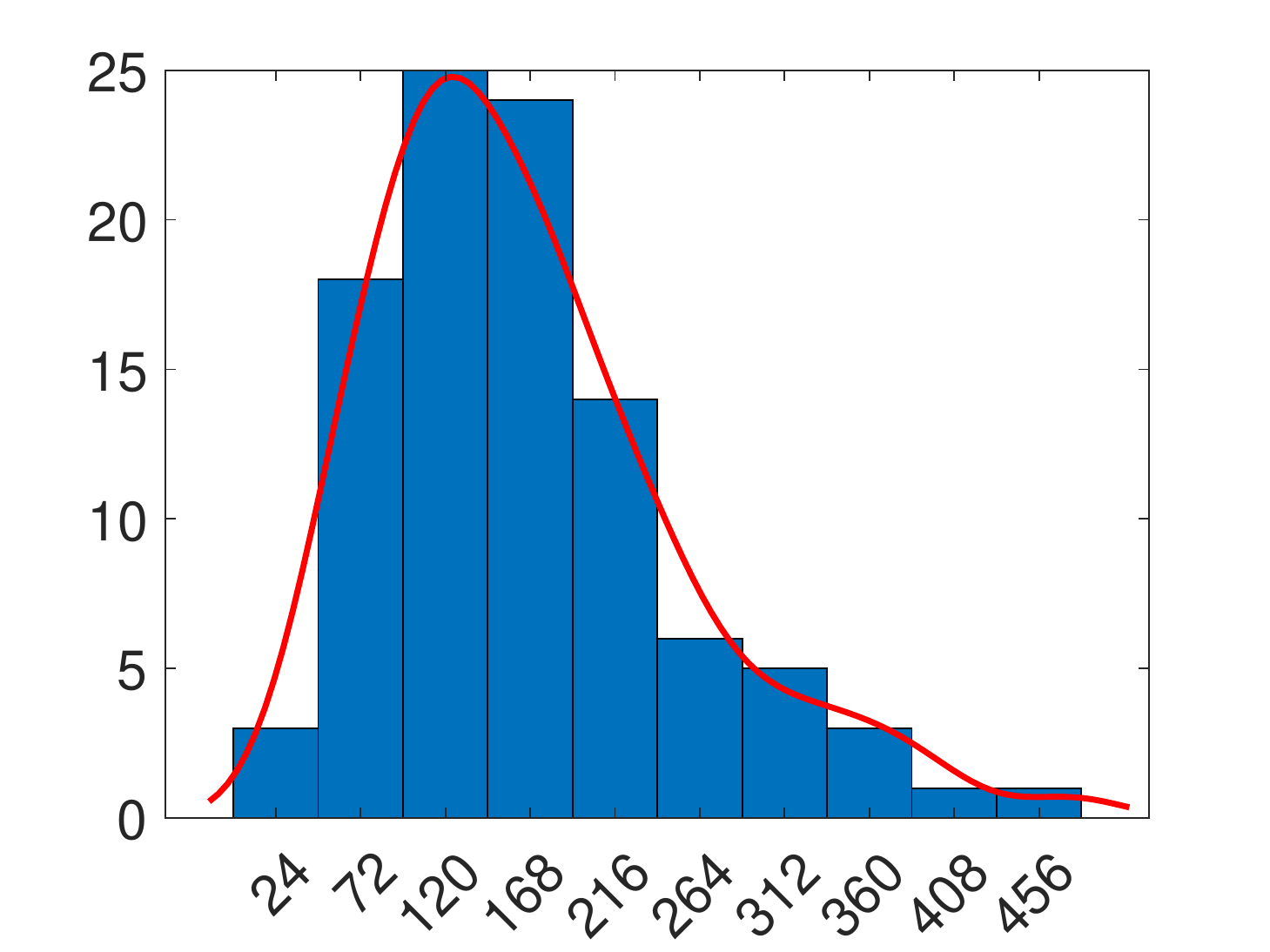}}$ & $\raisebox{-.5\height}{\includegraphics[scale=0.15]{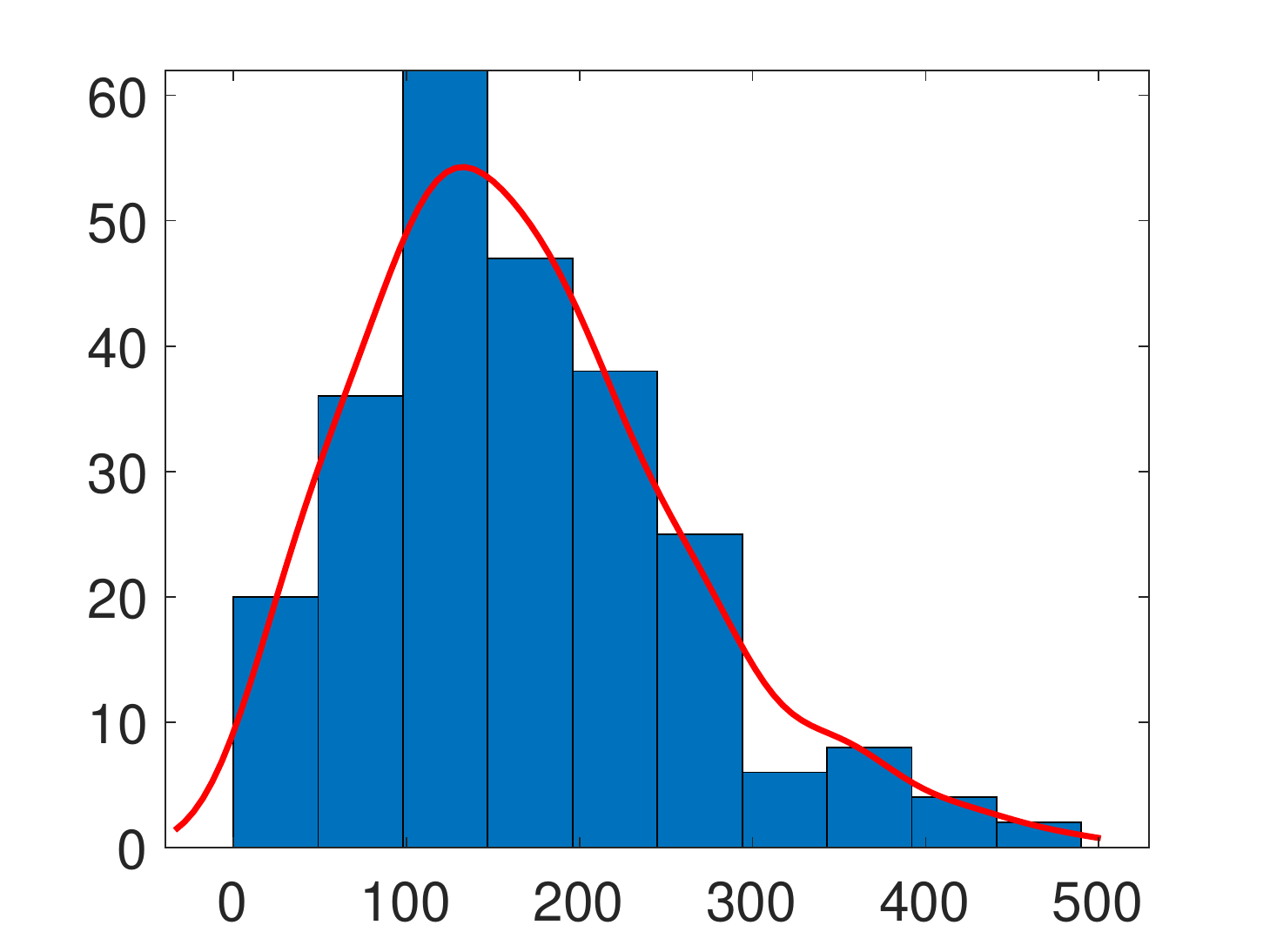}}$ \\
\hline
\hline
\end{tabular}
\end{threeparttable}
\end{center}
\label{Table2}
\end{table}

\subsubsection{Piece-wise Data Normalization}
\begin{figure}[H]
\centering
\subfigure[]{
\begin{minipage}[t]{1\linewidth}
\centering
\includegraphics[width=8.5cm]{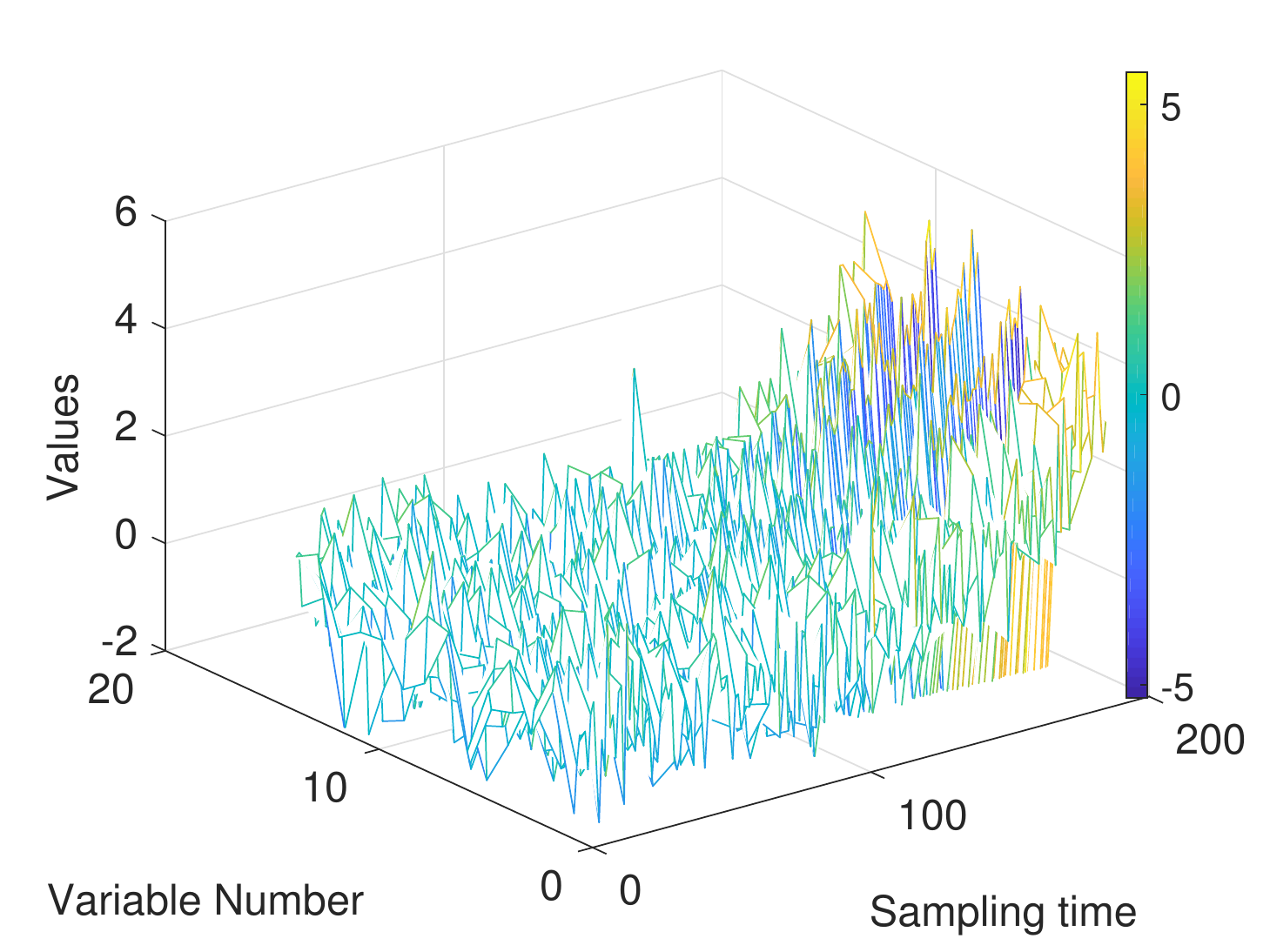}
\end{minipage}
}

\subfigure[]{
\begin{minipage}[t]{1\linewidth}
\centering
\includegraphics[width=8.5cm]{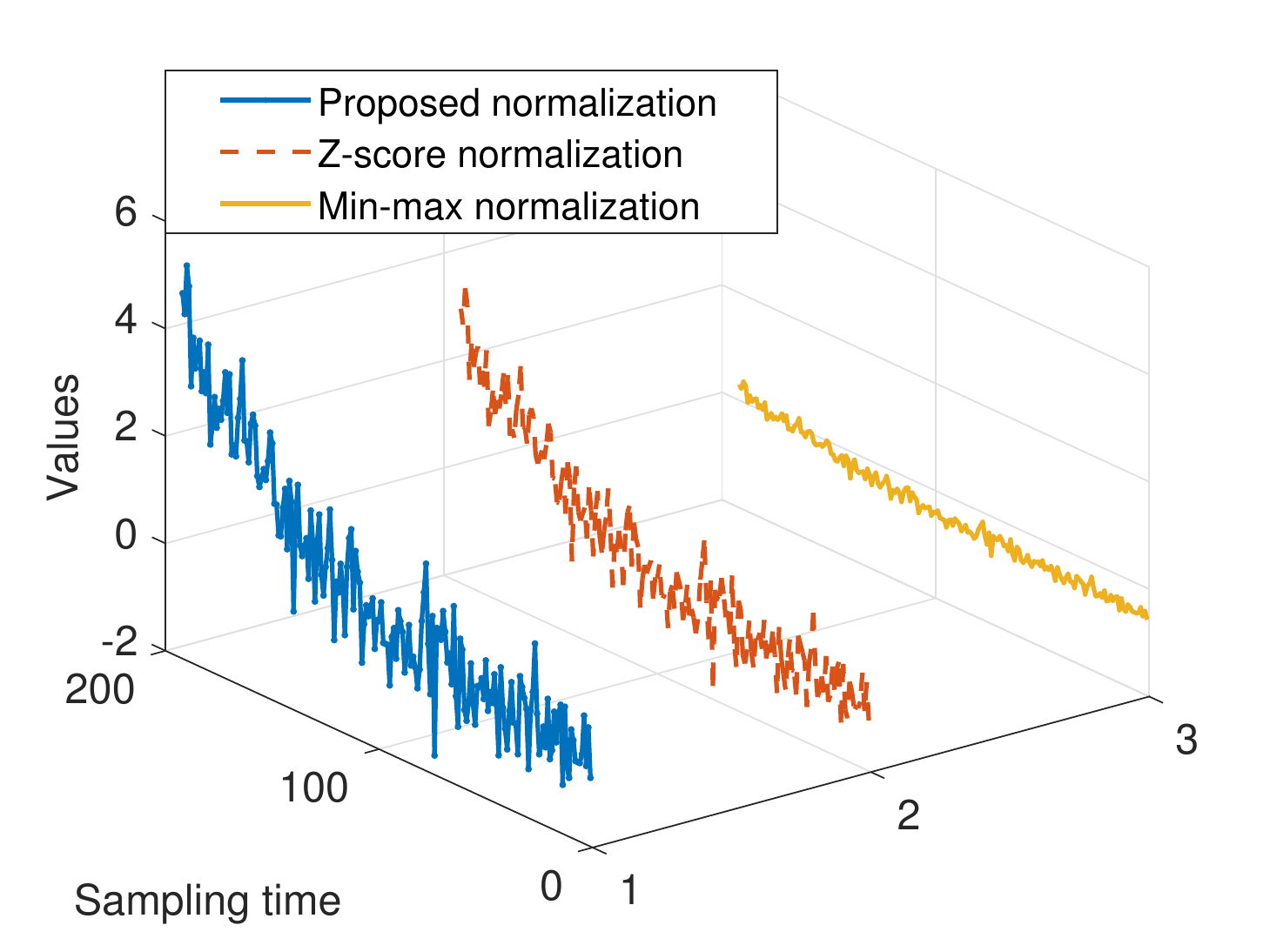}
\end{minipage}
}
\centering
\caption{(a) Normalized data of the first engine from FD001 using the proposed method and (b) Comparison between the proposed method and z-score and min-max normalization with the first variable of the first engine from FD001.}
\label{FIG6}
\end{figure}

We take the data from FD001 as an example to explain the ways of data processing. It is worth noticing that variables with constant values are removed from every trajectory to avoid disrupting estimation performance, and finally, the number of 14 variables is retained for analysis. Each trajectory in training data is divided into two parts with the given maximum RUL value, which is selected as 125 to keep consistency with most literature \cite{Ref12}. The data matrix $\mathbf X_n(10000 \times 14)$ is constructed by stacking all normal samples of each trajectory. With the normalization information, the degradation prat of each trajectory is normalized via Eq. (2). Fig. \ref{FIG6}(a) illustrates the normalized data of the first engine. Further, Fig. \ref{FIG6}(b) compares the first variable normalized by the proposed piece-wise normalization with another two standard methods. It is recognized that the proposed method amplifies the minor difference at the initial degradation stage and enlarges the difference between the beginning and the ending stage.

\subsubsection{Slow Feature Representation}
\begin{figure}[H]
\subfigure[]
{
\begin{minipage}[t]{0.45\linewidth}
\centering
\includegraphics[width=8cm]{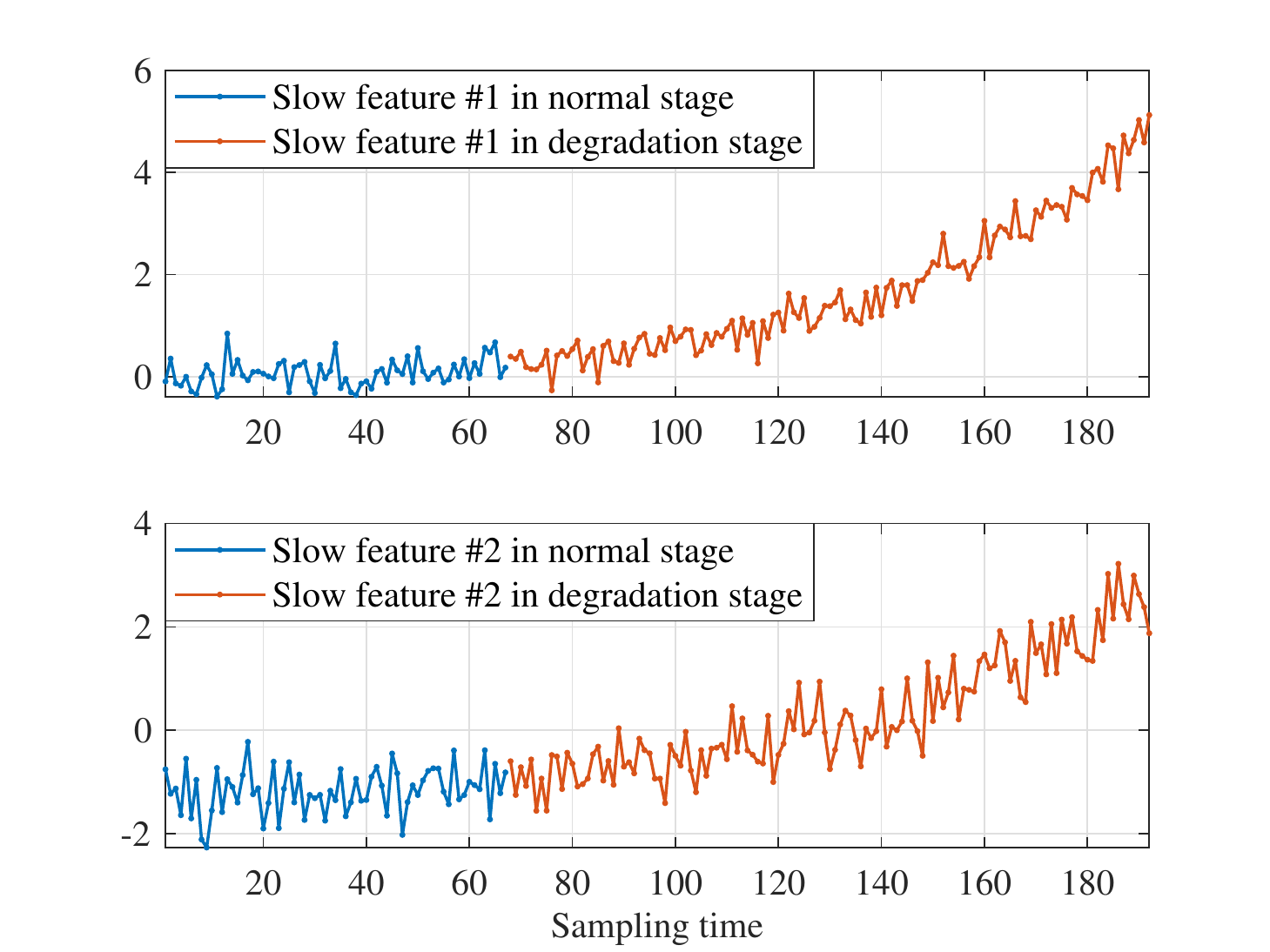}
\end{minipage}
}
\subfigure[]
{
\begin{minipage}[t]{0.45\linewidth}
\centering
\includegraphics[width=8cm]{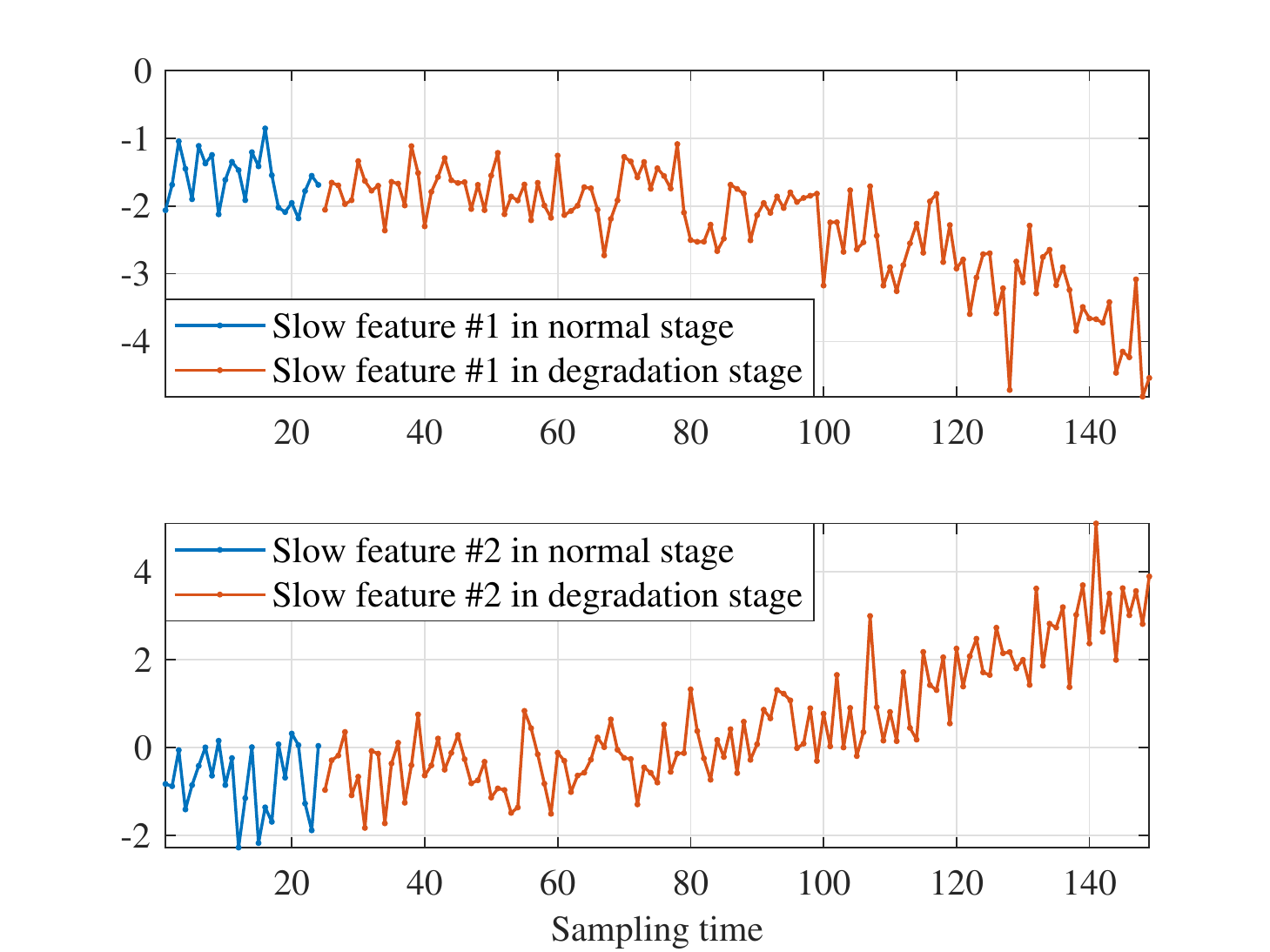}
\end{minipage}
}
\hfill
\subfigure[]
{
\begin{minipage}[t]{0.45\linewidth}
\centering
\includegraphics[width=8cm]{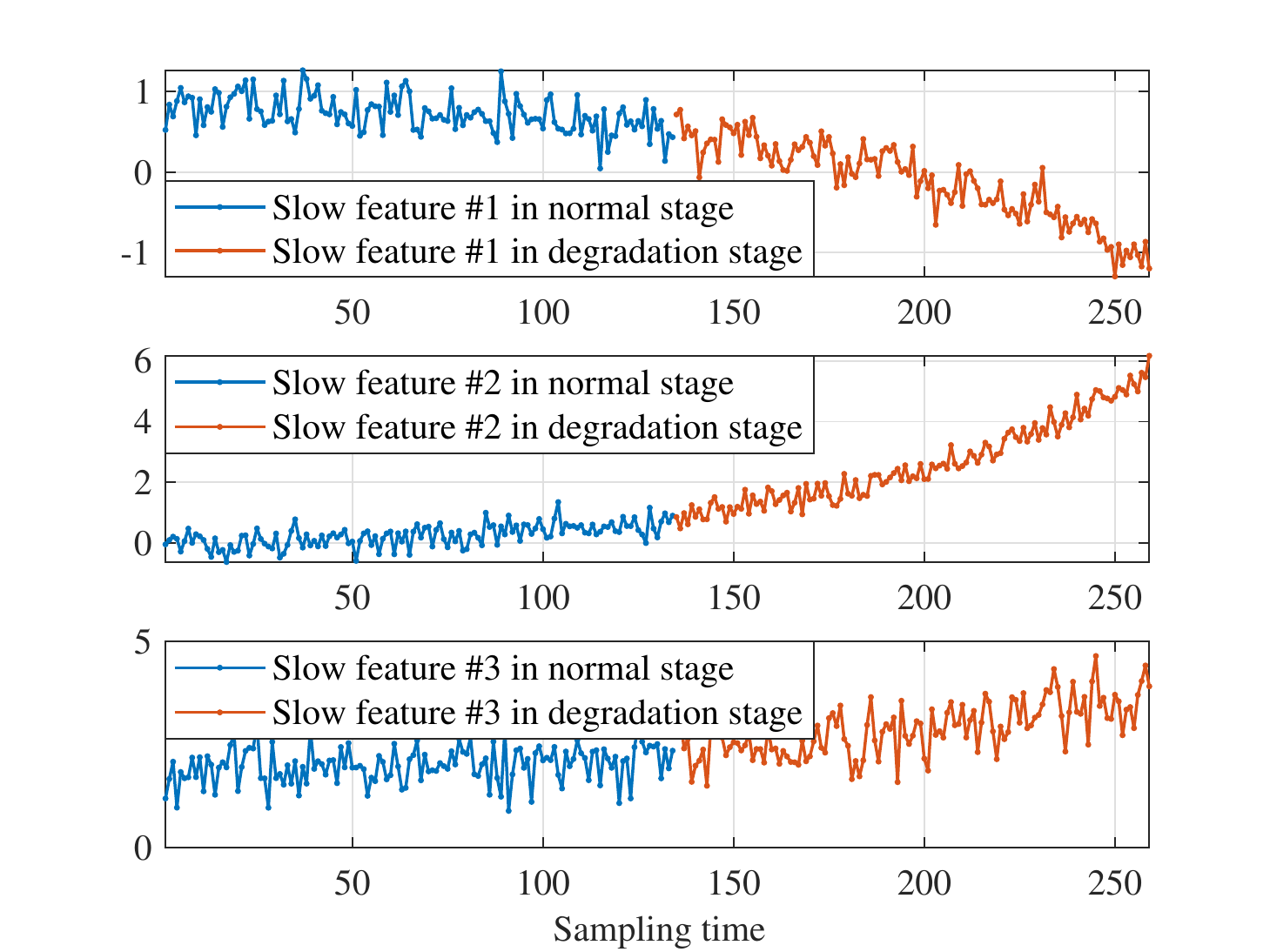}
\end{minipage}
}
\subfigure[]
{
\begin{minipage}[t]{0.45\linewidth}
\centering
\includegraphics[width=8cm]{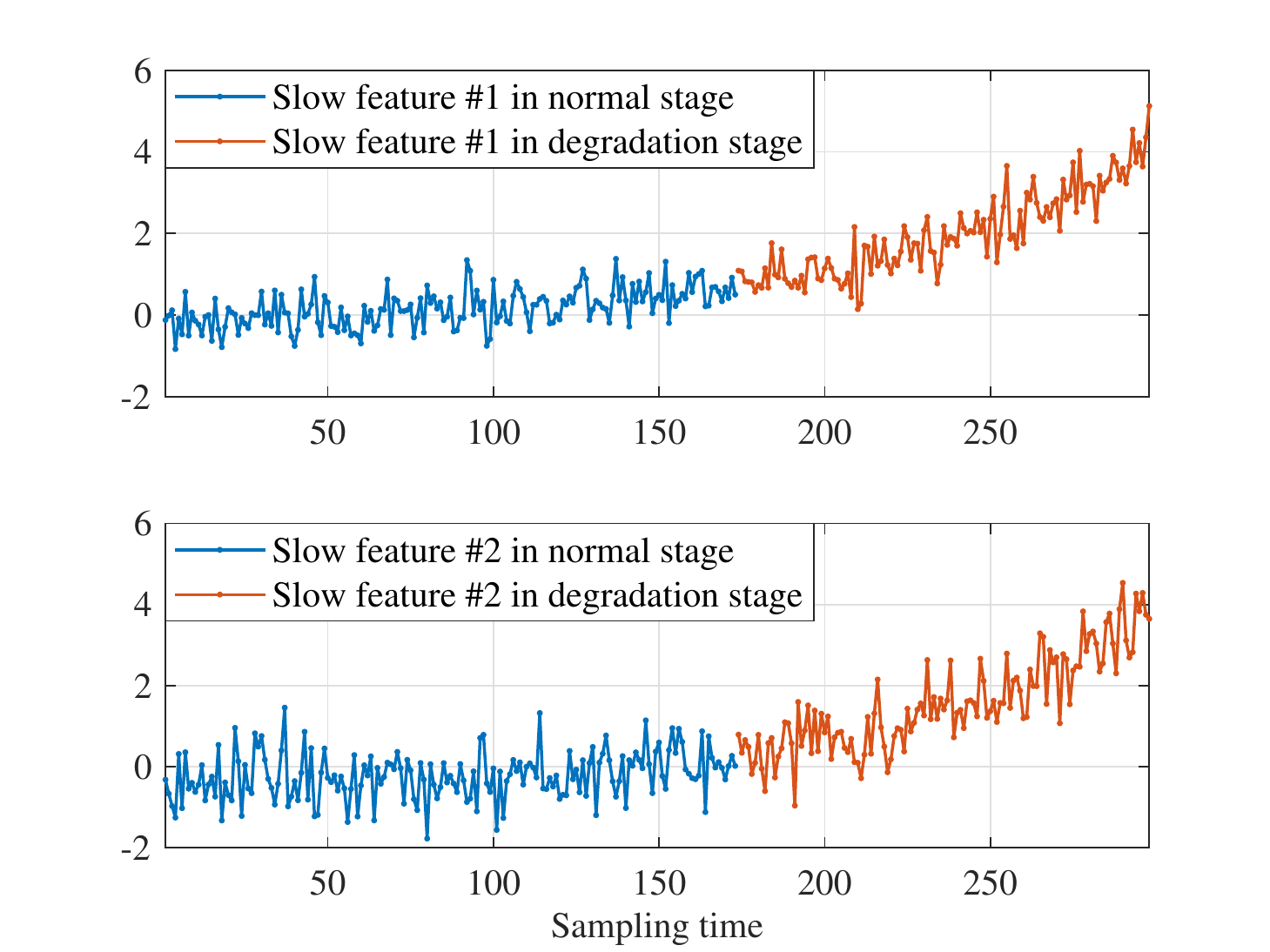}
\end{minipage}
}
\centering
\caption{The extracted slow features of Engine 1 in training data from (a)  FD001, (b) FD002, (c) FD003, and (d) FD004 (The blue line indicates the normal stage and the orange line stands for the fault stage).}
\label{FIG7}
\end{figure}

With the normal data $\mathbf X_n$, slow-varying subspace $\mathbf W_p$ is derived using Eqs. (4) and (5). The slow features with the smallest slowness are selected for sequential analysis by ranking each decomposed component with descending slowness. They are computed by projecting the normalized matrix $\mathbf X_{d,c}$ on $\mathbf W_p$. Fig. \ref{FIG7} visualizes the slow features of the first engine of each dataset. It is observed that slow features change slowly during the normal stage and then continually increase or decrease influenced by faults, revealing high sensitivity to the fault evolution. In contrast, the remaining residual features are plotted in Fig. 8. It is observed that all features vary rapidly and show almost consistent behaviour with zero mean and unit variance in both normal and degradation stages. It is reasonable to infer that all residual features are insensitive to fault evolution.

\begin{figure}[H]
\centering
\scriptsize
\includegraphics[scale=0.9]{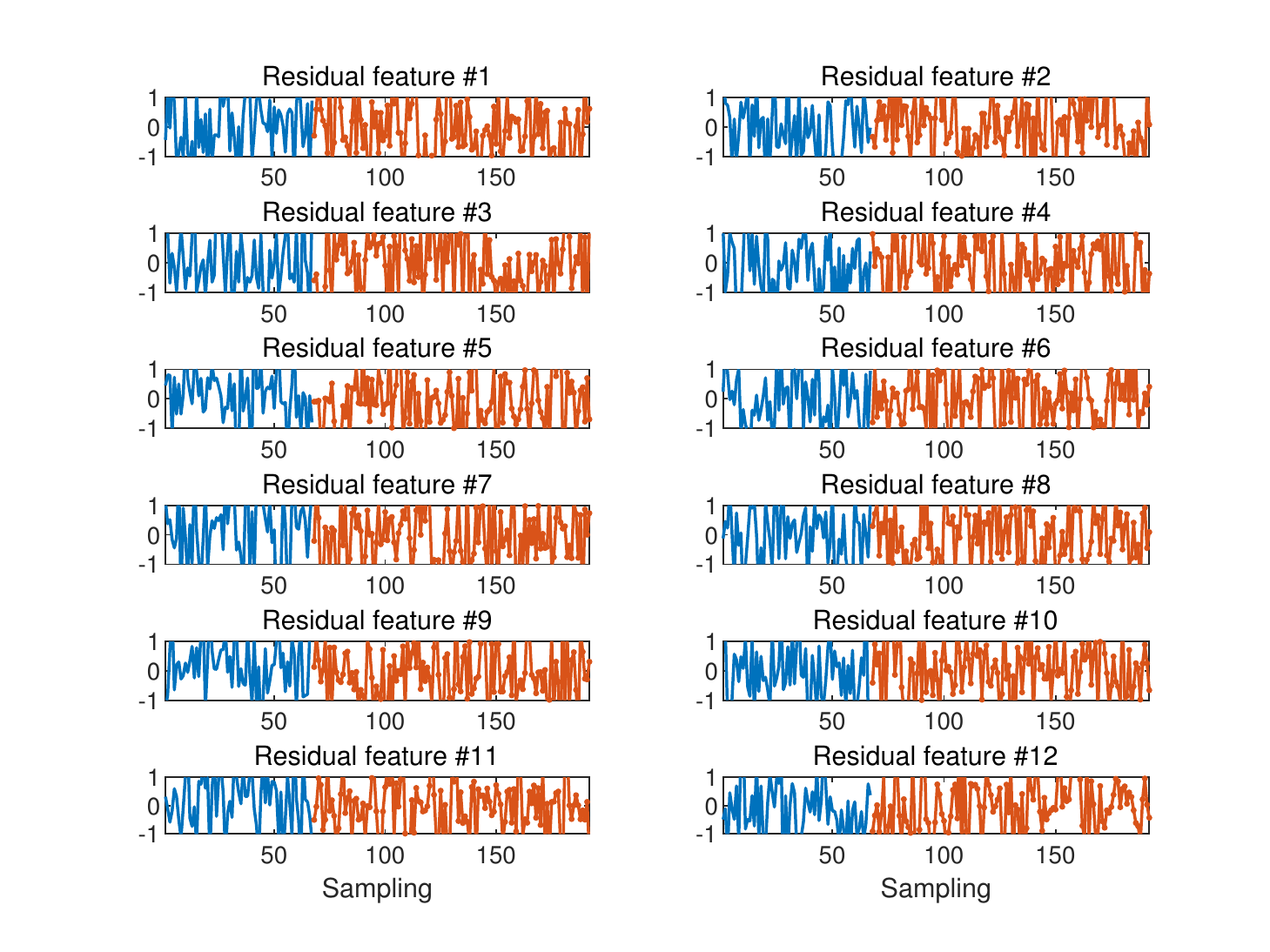} \\
\begin{center}
\caption{The features extracted from the residual subspace of Engine 1 in training data of FD001 (The blue line indicates the normal stage and the orange line stands for the fault stage).}
\end{center}
\label{FIG8}
\end{figure}

To further verify the above findings on a large time scale, Fig. 9 visualizes the distributions of residual features considering all training engines in FD001. As shown in the diagonal subfigures of Fig. 9, the distribution of residual features is typical Gaussian distributions with a zero mean and unit variation, revealing that faults make no difference on these features. Moreover, the scatter subplots enable the evaluation of the correlation between every two residual features in turn. For instance, subplots in the first row measure the correlation between the first feature and the remaining ones sequentially. In most cases, the distributions are almost sphering, indicating they are still orthogonal even within the faulty stage. Similar observations could be concluded by applying the aforementioned procedures to the other three datasets, i.e., FD002, FD003, and FD004.

\begin{figure}[H]
\centering
\includegraphics[scale=0.8]{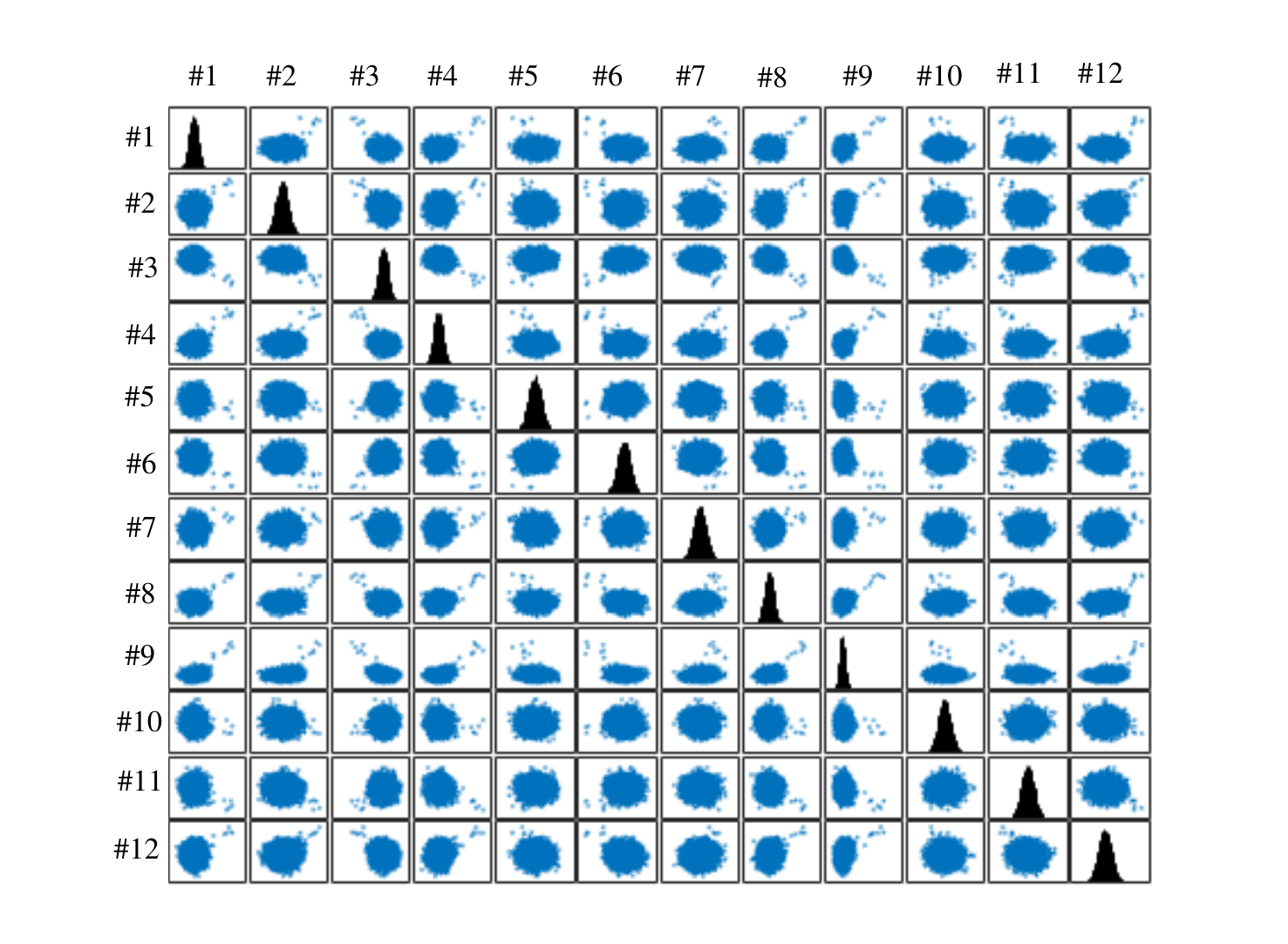} \\
\begin{center}
\caption{The data distribution of residual features and their correlations regarding all training engines from FD001.}
\end{center}
\label{FIG9}
\end{figure}

The obtained slow features $\mathbf S_{d,c}$ are stacked with the normalized $\mathbf X_{d,c}$ to construct the hybrid features $\mathbf T_c$. In \cite{Ref15}, sliding window length $L$ is recommended as 30 for FD001 and 50 for FD004 through trial and error. Contrarily, we adopt sample auto-correlation theory to accurately determine the value for $L$ using extracted slow feature, as shown in Fig. 10. Here, we use the first slow feature to determine $L$, which is 28 for FD001. Performing similar steps for the other three datasets, the corresponding hybrid datasets are arranged. The value of $L$ for FD002, FD003, and FD004 are 60, 56, and 48, respectively.

\begin{figure}[H]
\subfigure[]
{
\begin{minipage}[t]{0.45\linewidth}
\centering
\includegraphics[width=6cm]{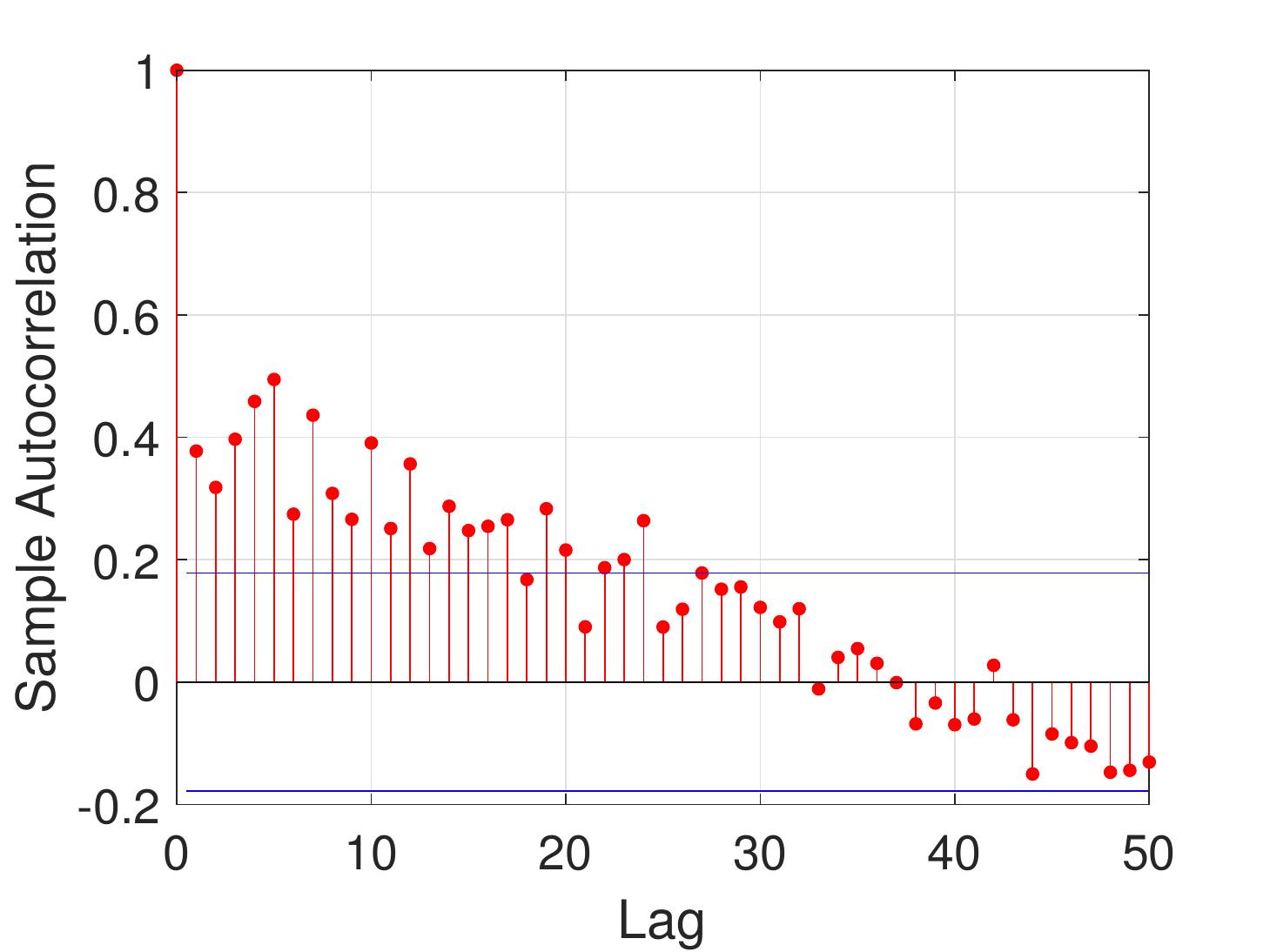}
\end{minipage}
}
\subfigure[]
{
\begin{minipage}[t]{0.45\linewidth}
\centering
\includegraphics[width=6cm]{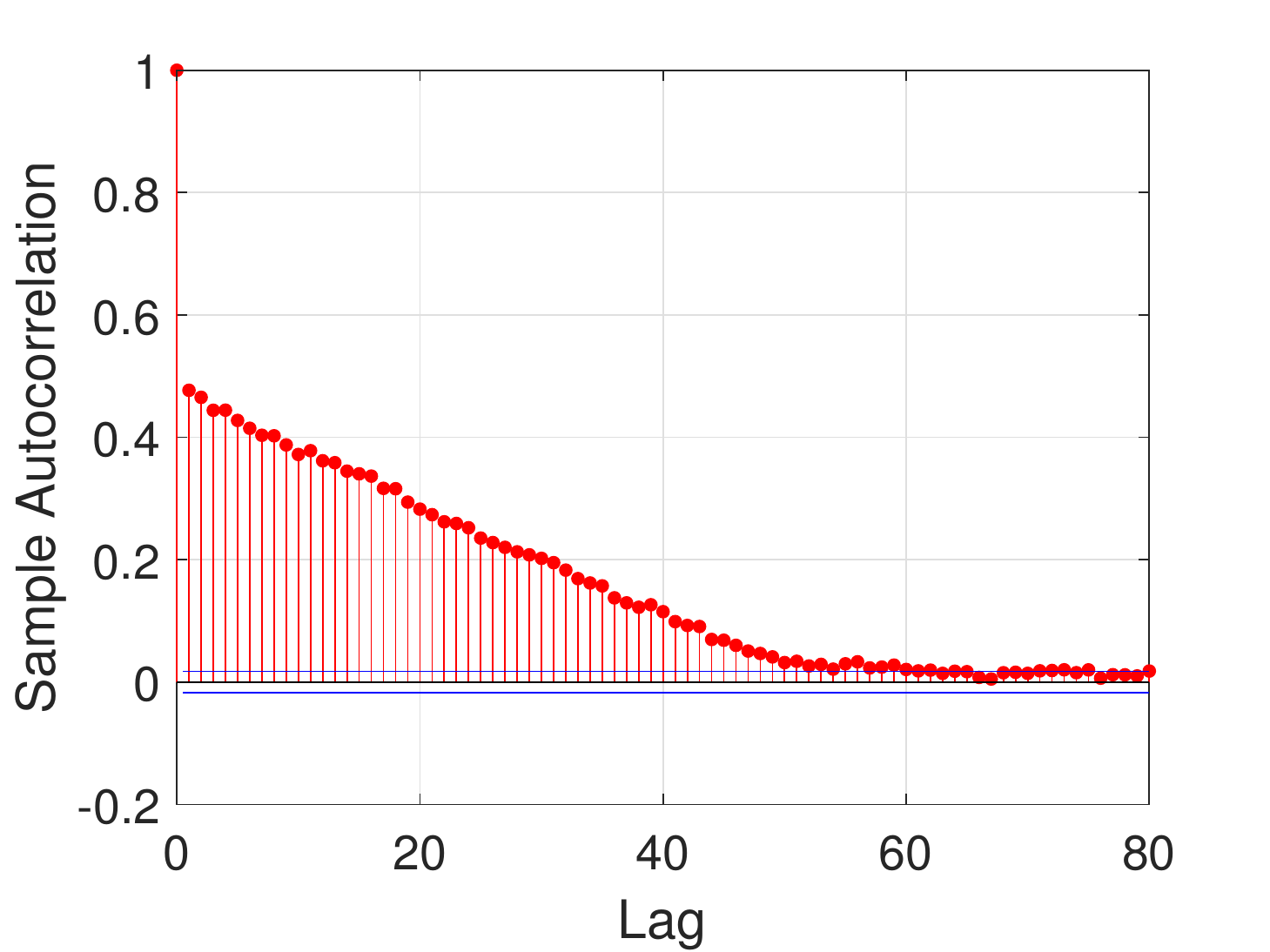}
\end{minipage}
}
\hfill
\subfigure[]
{
\begin{minipage}[t]{0.45\linewidth}
\centering
\includegraphics[width=6cm]{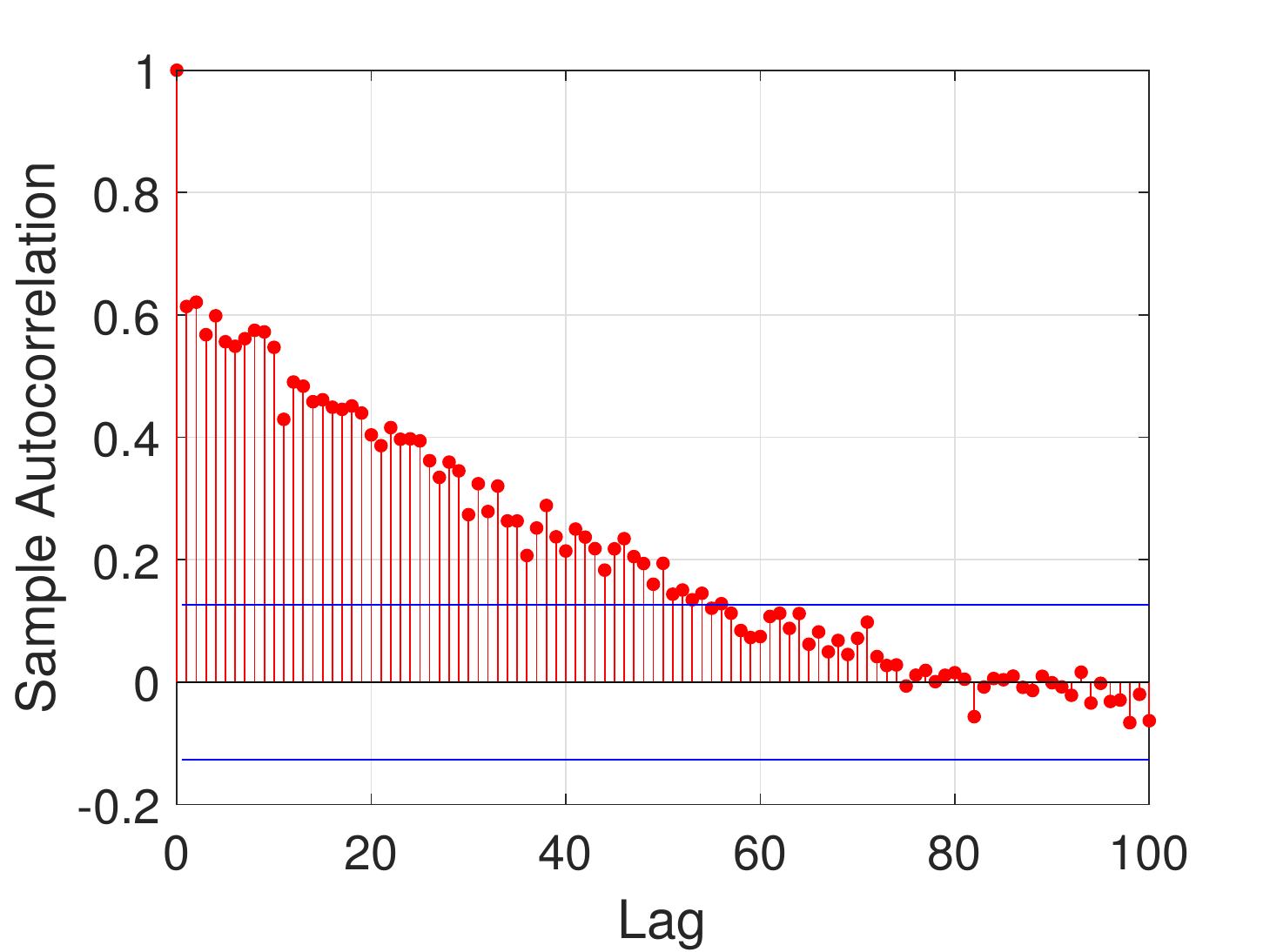}
\end{minipage}
}
\subfigure[]
{
\begin{minipage}[t]{0.45\linewidth}
\centering
\includegraphics[width=6cm]{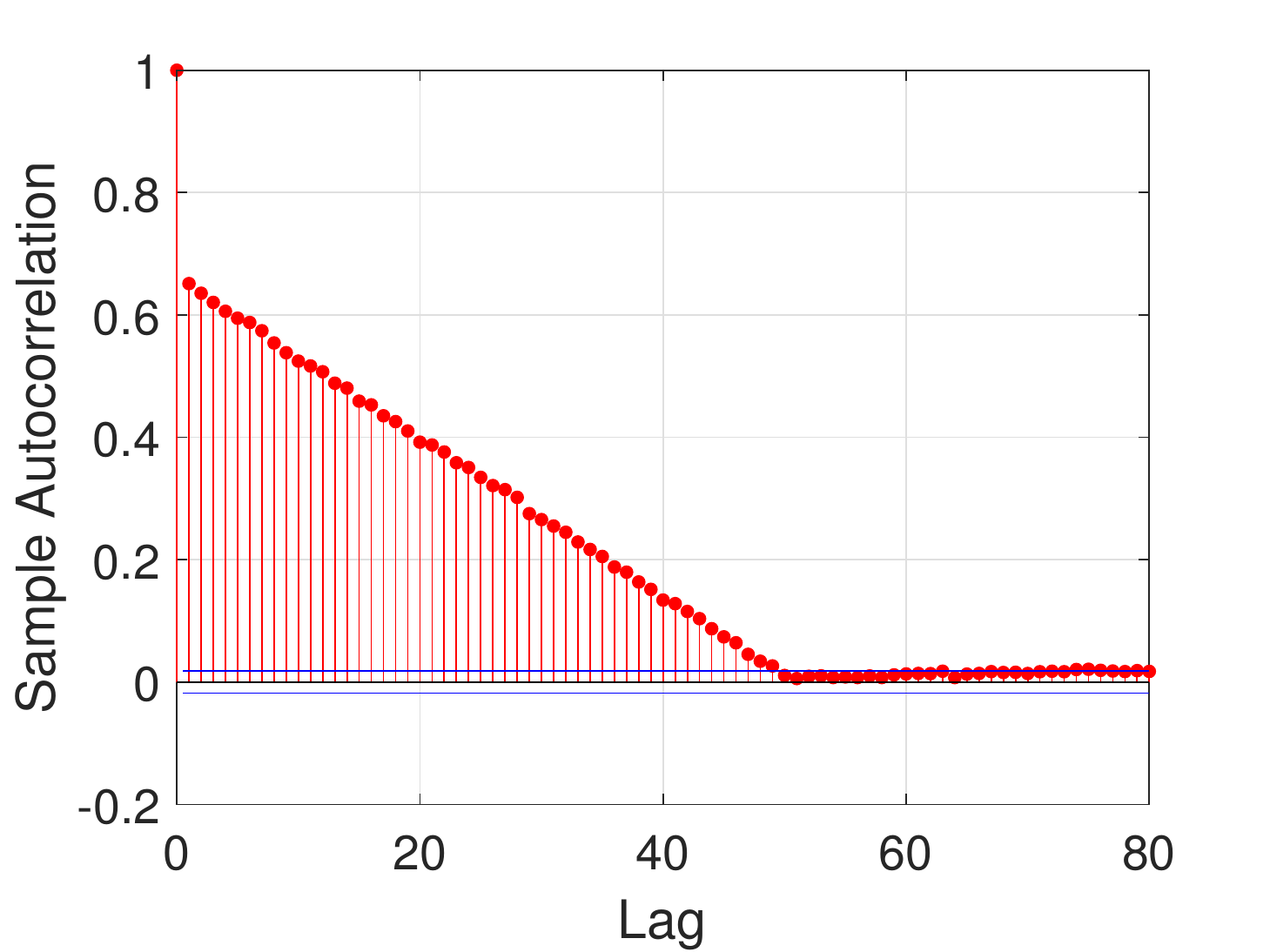}
\end{minipage}
}
\centering
\caption{Selection of window length using the extracted slow feature for (a) FD001, (b) FD002, (c) FD003, and (d) FD004.}
\label{Fig9}
\end{figure}

\begin{figure}[H]
\subfigure[]
{
\begin{minipage}[t]{0.45\linewidth}
\centering
\includegraphics[width=6cm]{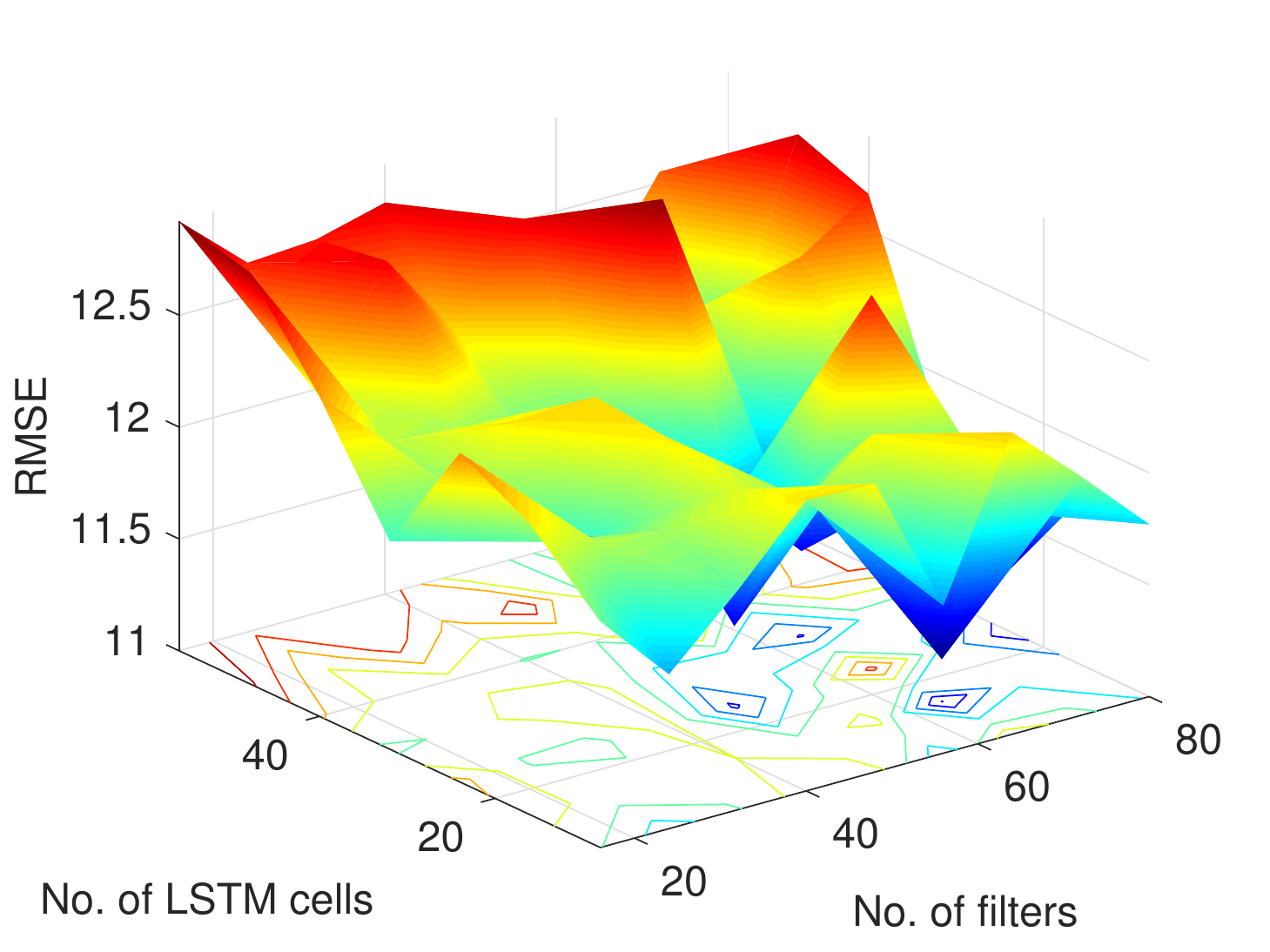}
\end{minipage}
}
\subfigure[]
{
\begin{minipage}[t]{0.45\linewidth}
\centering
\includegraphics[width=6cm]{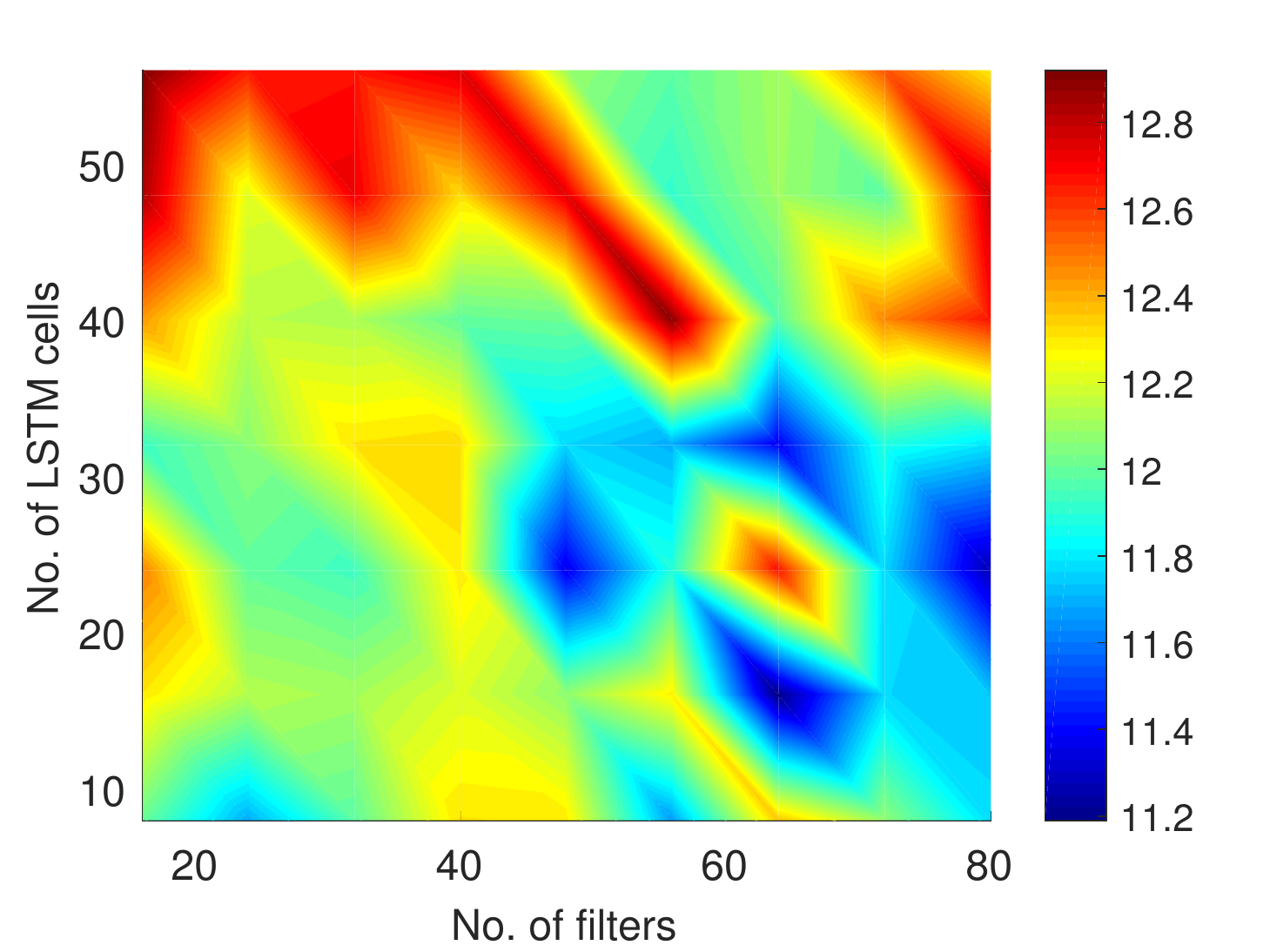}
\end{minipage}
}
\hfill
\subfigure[]
{
\begin{minipage}[t]{0.45\linewidth}
\centering
\includegraphics[width=6cm]{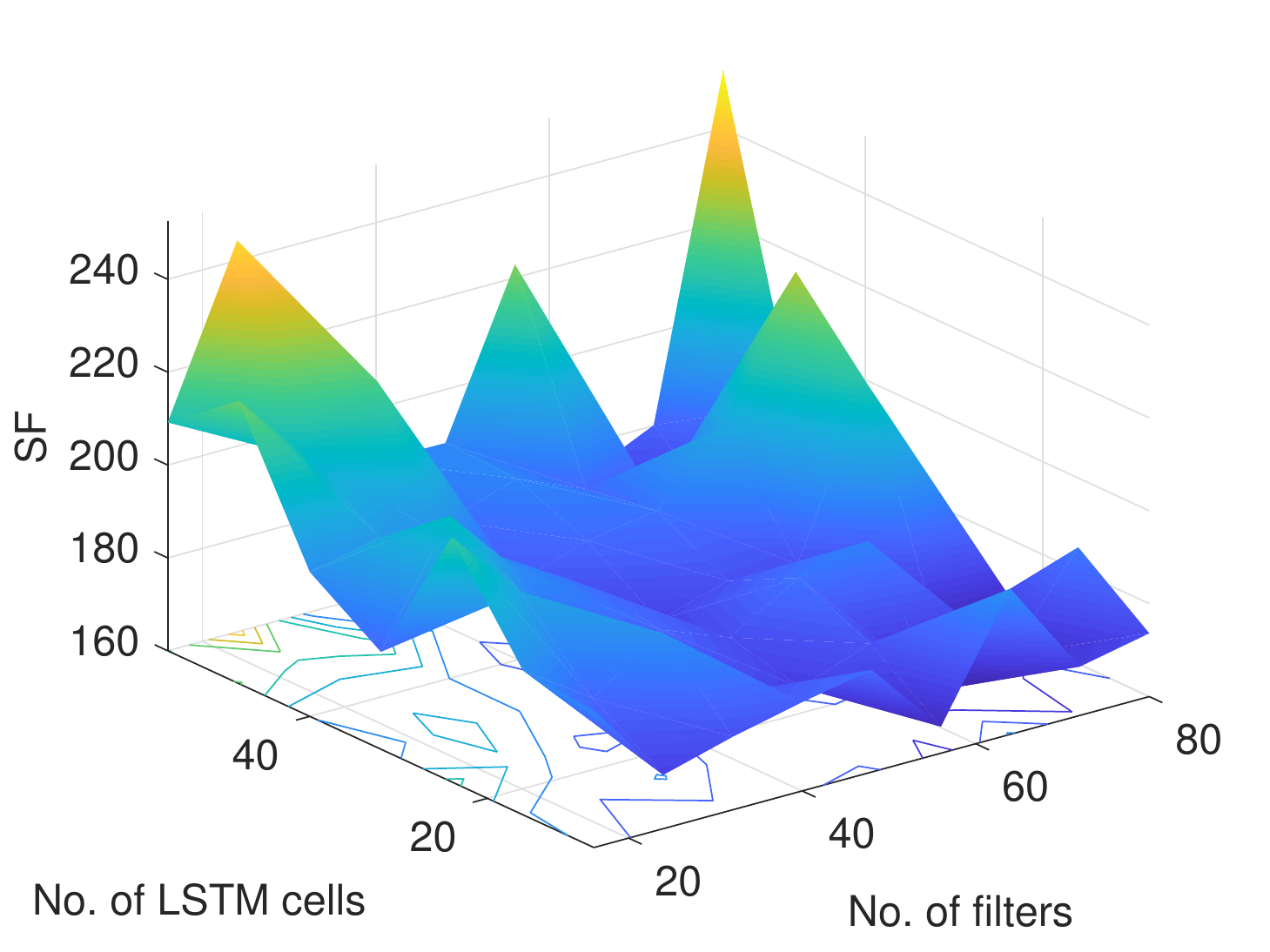}
\end{minipage}
}
\subfigure[]
{
\begin{minipage}[t]{0.45\linewidth}
\centering
\includegraphics[width=6cm]{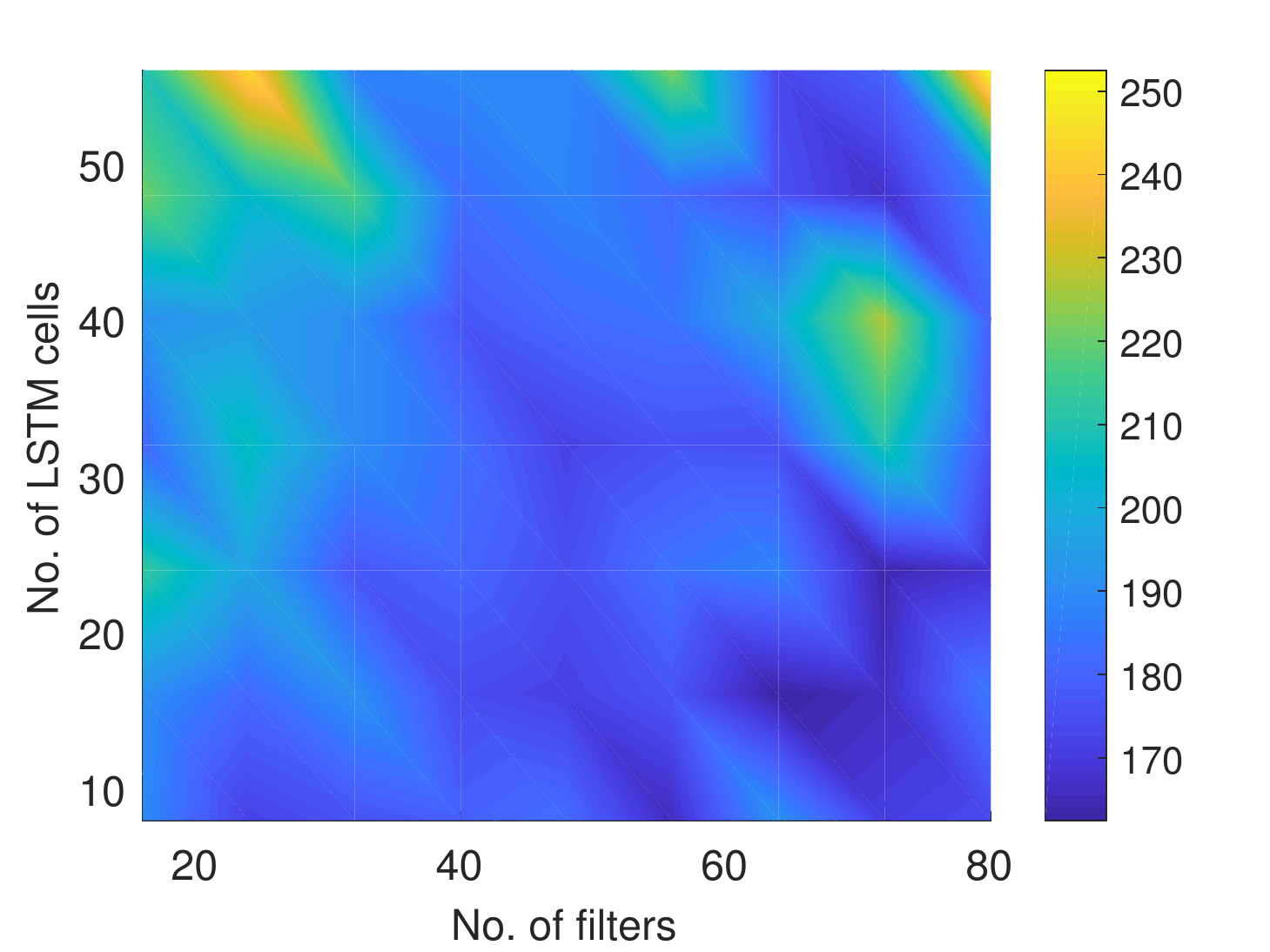}
\end{minipage}
}
\centering
\caption{Sensitive analysis of crucial parameters for FD001 with respect to (a) $RMSE$ in three-dimensional view, (b) $RMSE$ in plain view, (c) $SF$ in three-dimensional view, and (d) $SF$ in plain view.}
\label{Fig11}
\end{figure}

\subsubsection{Network Configuration}
For clear understanding, FD001 is still taking as an example to illustrate the adjustment of tuneable parameters in SD-TemCapsNet. First, the dimension of advanced capsules is given to 16, as the same value of the number of variables in a data frame. Then the number of advanced capsules is the same as the number of slow features, which is 2 here. Correspondingly, the dimension of basic capsules is calculated as 8. The number of LSTM cells and the number of filters are searched starting from 8 with the interval 8. Fig. 11(a) illustrates the values of $RMSE$ in each combination. From the plain view given in Fig. 11(b), the specific value of LSTM cells and filter values corresponding to the best estimation accuracy are 16 and 64, respectively. Similar conclusions could be drawn with index $SF$, as shown in Figs. 11(c) and 11(d).

The specific network configurations concerning all four datasets are summarized in Table III. Although operating conditions and fault types of these four datasets differ from each other, the networks share the same values for most parameters. Specifically, the solution to determine window length has been addressed in the last subsection. The number of epochs can be found through early stopping. From Fig. 12(a), SD-TemCapsNet contains more parameters than other models. Therefore, more training time is naturally assumed to be required, as shown in Fig. 12(b). However, with the assist of slow-varying dynamics, much parameter tuning time is saved by avoiding extensive trial and error.

\begin{table}[H]
\renewcommand{\arraystretch}{1.5}
\centering
\caption{Network configurations of SD-TemCapsNet for all data sets collected from engine system.}
\scriptsize
\begin{tabular}{c c c c c c}
\hline
\hline
\multirow{2}{*}{\textbf{Layer Name}} & \multirow{2}{*}{\textbf{Parameter Name}}  & \multicolumn{4}{c}{\textbf{Dataset Name}}\\
\cline{3-6}
&  & {FD001} & {FD002} & {FD003} & {FD004} \\
\hline
\multirow{2}{*}{Input} & {Epoch}  & 80 & 40 & 80 & 40 \\
\cline{2-6}
& {Window length}  & 28 & 60 & 56 & 48 \\
\hline
\multirow{3}{*}{Convolution} & {Filters}  & 64 & 64 & 32 & 64 \\
\cline{2-6}
& {Kernal size}  & (1,2) & (1,2) & (1,2) & (1,2) \\
\cline{2-6}
& {Strides}  & (1,2) & (1,2) & (1,2) & (1,2) \\
\hline
\multirow{4}{*}{Basic capsule} & {Dimensions}  & 8 & 8 & 4 & 8 \\
\cline{2-6}
& {Channels} & 8 & 8 & 8 & 8 \\
\cline{2-6}
& {Kernal size} & (1,8) & (1,8) & (1,8) & (1,8) \\
\cline{2-6}
& {Strides}  & (1,1) & (1,1) & (1,1) & (1,1) \\
\hline
\multirow{2}{*}{Advanced capsule} & {No. of Capsules}  & 2 & 2 & 3 & 2 \\
\cline{2-6}
& {Dimensions}  & 16 & 16 & 17 & 16 \\
\hline
{LSTM} & {Units}  & 16 & 32 & 16 & 32 \\
\hline
\multirow{3}{*}{Output} & {The first FNN} & \multicolumn{4}{c}{No. of Neurons=200, Dropout ratio=0.2} \\
& {The second FNN}  & \multicolumn{4}{c}{No. Neurons=100, Dropout ratio=0.2} \\
& {The third FNN}      & \multicolumn{4}{c}{No. Neurons=1} \\
\hline
\hline
\end{tabular}
\end{table}

\begin{figure}[H]
\setcounter{subfigure}{0}
\subfigure[]
{
\begin{minipage}[t]{1\linewidth}
\centering
\includegraphics[width=7cm]{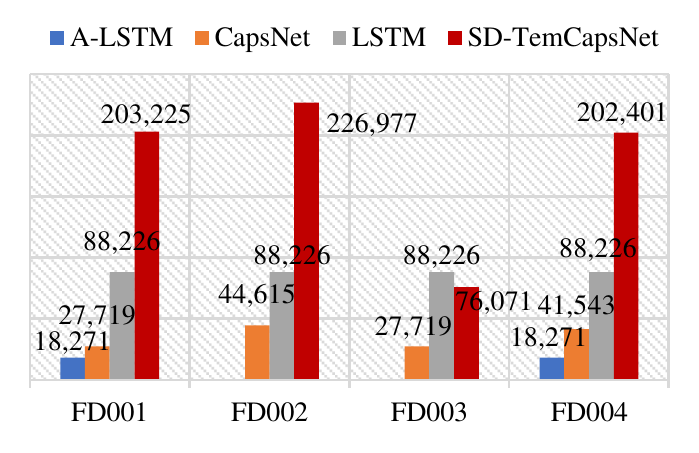}
\end{minipage}
}

\subfigure[]
{
\begin{minipage}[t]{1\linewidth}
\centering
\includegraphics[width=7cm]{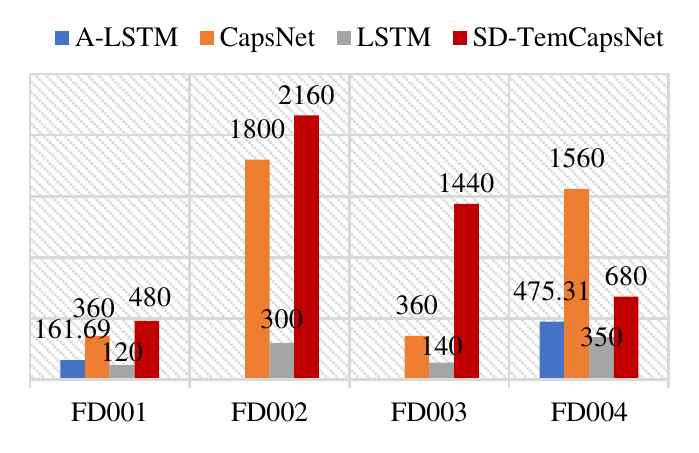}
\end{minipage}
}
\centering

\caption{Comparison between SD-TemCapsNet and the selected methods regarding (a) total parameter amount and (b) training time (second).}
\end{figure}

\subsubsection{Performance Comparison}
With the configuration given in Table III, the proposed SD-TemCapsNet has been performed on each dataset over ten runs, contributing to evaluating robustness. Table IV compares SD-TemCapsNet with seven popular DNN-based RUL estimation models concerning two indices, i.e., $RMSE$ and $SF$. For FD001, Deep CNN, S-DNN, and CapsNet present similar estimation errors around 12.56 with $RMSE$, but they are much better than the other four methods. The proposed method is capable of significantly reducing the value of $RMSE$ to 11.19 and the value of $SF$ to 162.49, achieving the best results so far. In other words, accuracy improvements concerning $RMSE$ and $SF$ are gained by 10.17$\%$ and 34.78$\%$, respectively. The improvement is computed by the ratio between reduced and original errors, e.g., (12.58-11.30)/12.58. The proposed method still outperforms its counterparts with noticeable improvement for the other datasets with more complex operating conditions.

\begin{table}[H]
\renewcommand{\arraystretch}{1.5}
\centering
\caption{Accuracy comparison between the proposed method over ten runs with indices $RMSE$ and $SF$ and the current methods.}
\tiny
\begin{tabular}{c|c|c|c|c|c|c|c|c|c|c}
\hline
\hline
\multirow{2}*{\textbf{Dataset}} & \multirow{2}*{\textbf{Index}}  & \multicolumn{7}{c|}{\textbf{The Existing Models}} & \multicolumn{2}{c}{\textbf{Proposed}} \\
\cline{3-11}
& & \textbf{CNN$^{[13]}$} & \textbf{LSTM$^{[15]}$} & \textbf{MODBNE$^{[29]}$} & \textbf{A-LSTM $^{[18]}$} & \textbf{Deep CNN$^{[14]}$} & \textbf{S-DNN$^{[30]}$} & \textbf{CapsNet$^{[12]}$} & \textbf{Accuracy} & \textbf{IMA} \\
\hline
\multirow{2}{*}{FD001} & $RMSE$ & 18.45 & 16.14 & 15.04 & 14.53 & 12.61±0.19 & 12.56 & 12.58±0.25 & \textbf{11.30±0.08} & 10.17$\%$\\
& $SF$ & 1286.70 & 338.00 & 334.00 & 322.44 & 273.7±24.1 & 231.00 & 276.34±25.95 & \textbf{180.23±9.63} & 34.78$\%$\\
\hline
\multirow{2}{*}{FD002} & $RMSE$ & 30.29 & 24.49 & 25.05 & NA & 22.36±0.32 & 22.73 & 16.30±0.23 & \textbf{12.23±0.34} & 24.97$\%$\\
& $SF$ & 13570.00 & 445.00 & 5585.00 & NA & 10412.00±544.00 & 3366.00 & 1229.72±53.07 & \textbf{360.52±23.38} & 70.68$\%$\\ 
\hline
\multirow{2}{*}{FD003} & $RMSE$ & 19.82 & 16.18 & 12.51 & NA & 12.64±0.14 & 12.10 & 11.71±0.26 & \textbf{11.33±0.21} & 3.25$\%$\\
& $SF$ & 1596.20 & 852.00 & 422.00 & NA & 284.1±26.5 & 251.00 & 283.81±29.46 & \textbf{176.31±9.17} & 37.88$\%$\\
\hline
\multirow{2}{*}{FD004} & $RMSE$ & 29.16 & 28.17 & 28.66 & 27.08 & 23.31±0.39 & 22.66 & 18.96±0.27 & \textbf{16.49±0.34} & 13.03$\%$\\ 
& $SF$ & 7886.40 & 5500.00 & 6558.00 & 5649.14 & 12466.00±853.00 & 2840.00 & 2625.64±266.83 & \textbf{804.05±37.44} & 69.38$\%$\\
\hline
\hline
\end{tabular}
\begin{tablenotes}
     \item[1] NA indicates the results are not provided in the cited work;
     \item[2] IMA is short for improved prediction accuracy (IMA) for average error compared to CapsNet.
\end{tablenotes}
\end{table}

Fig. 13 compares the prediction performance of the proposed method with that of the original CapsNet approach on testing engines in all datasets, which includes both prediction values and the corresponding prediction errors. Significantly, the error distributions presented in the area plot indicate that the proposed method yields better results, as they are totally covered by that of CapsNet.

\textit{Remark:} All competing algorithms used for comparison follow the same network configuration as reported, contributing to fair comparison.
\begin{figure}[H]
\subfigure[]
{
\begin{minipage}[t]{0.45\linewidth}
\centering
\includegraphics[width=6.5cm]{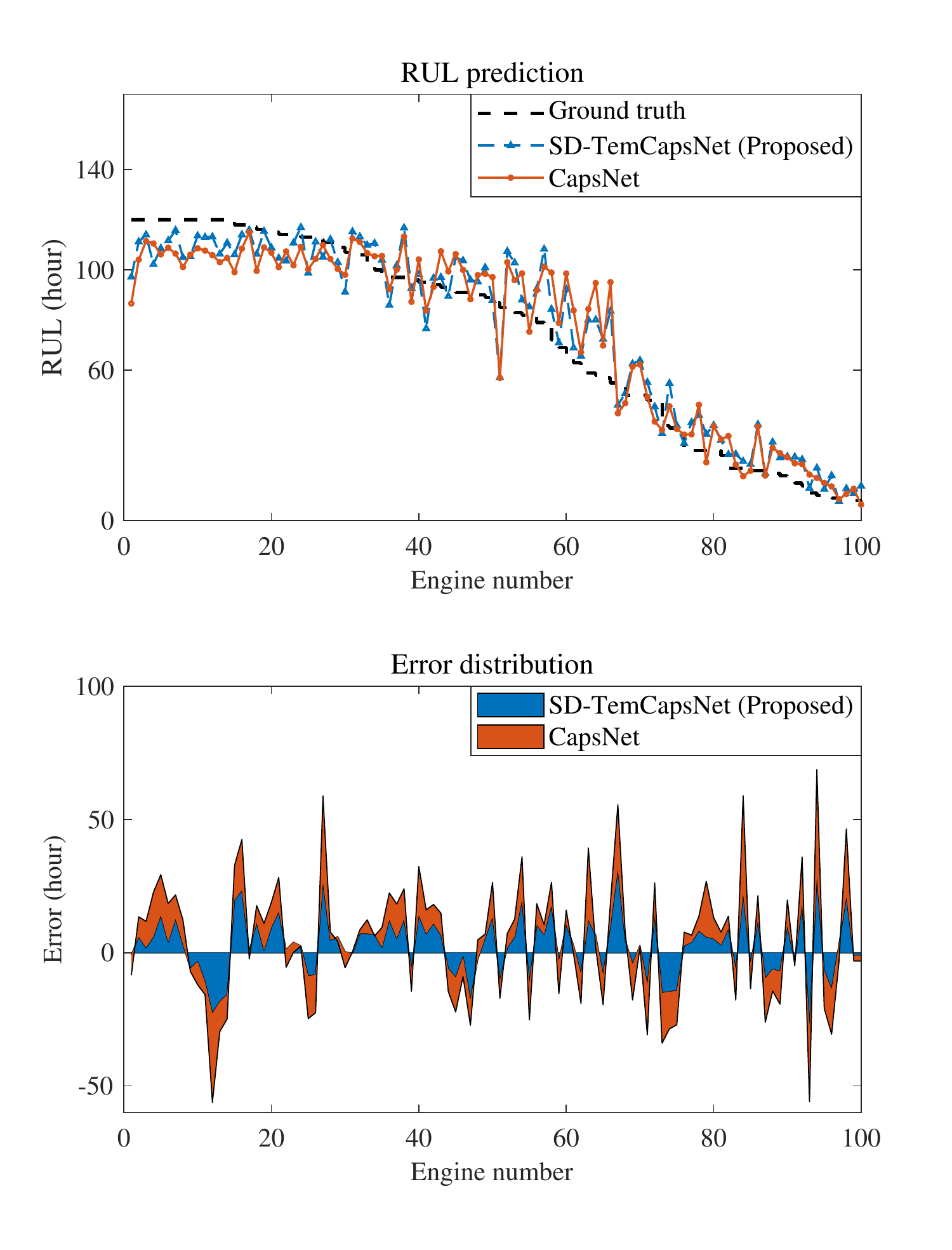}
\end{minipage}
}
\subfigure[]
{
\begin{minipage}[t]{0.45\linewidth}
\centering
\includegraphics[width=6.5cm]{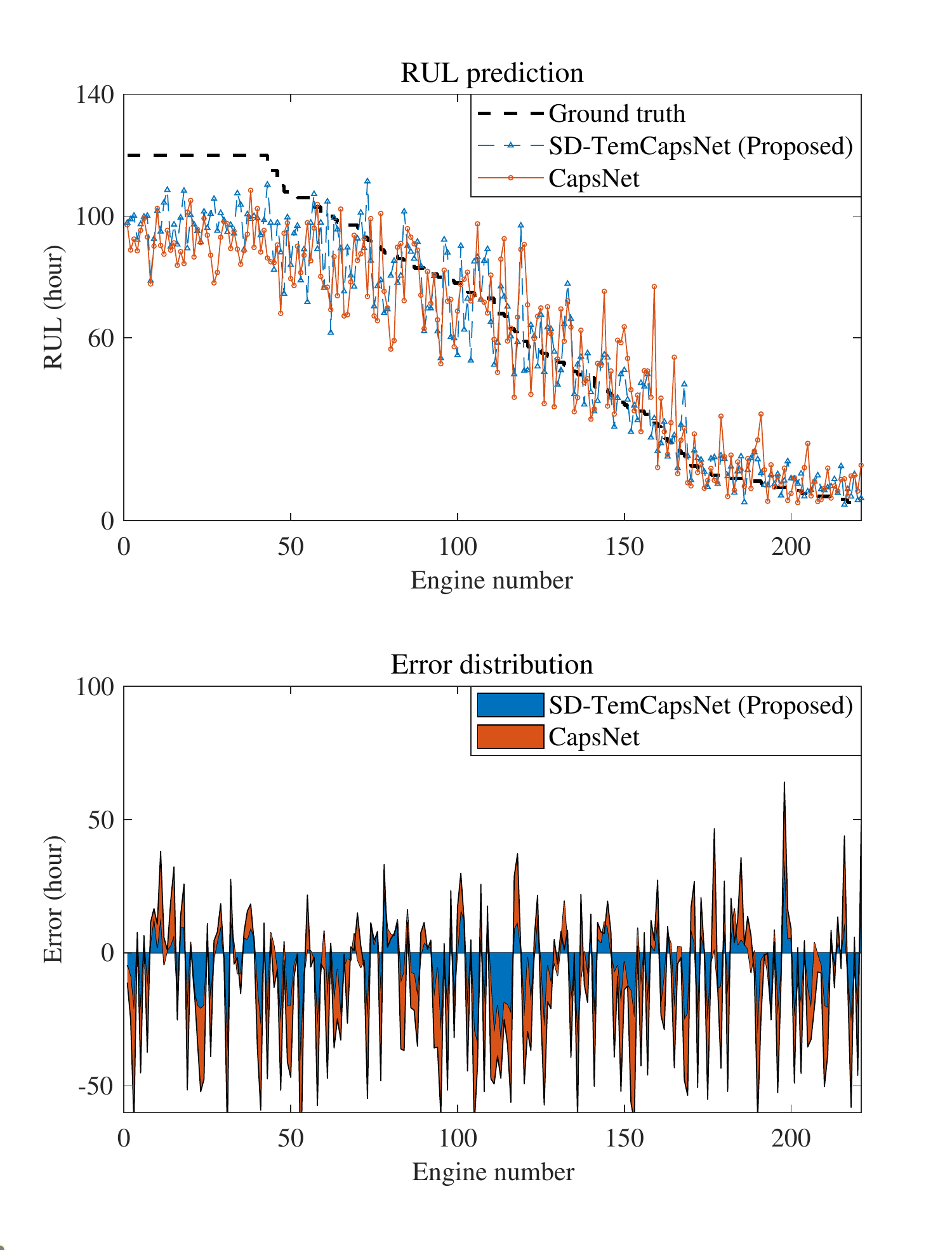}
\end{minipage}
}
\hfill
\subfigure[]
{
\begin{minipage}[t]{0.45\linewidth}
\centering
\includegraphics[width=6.5cm]{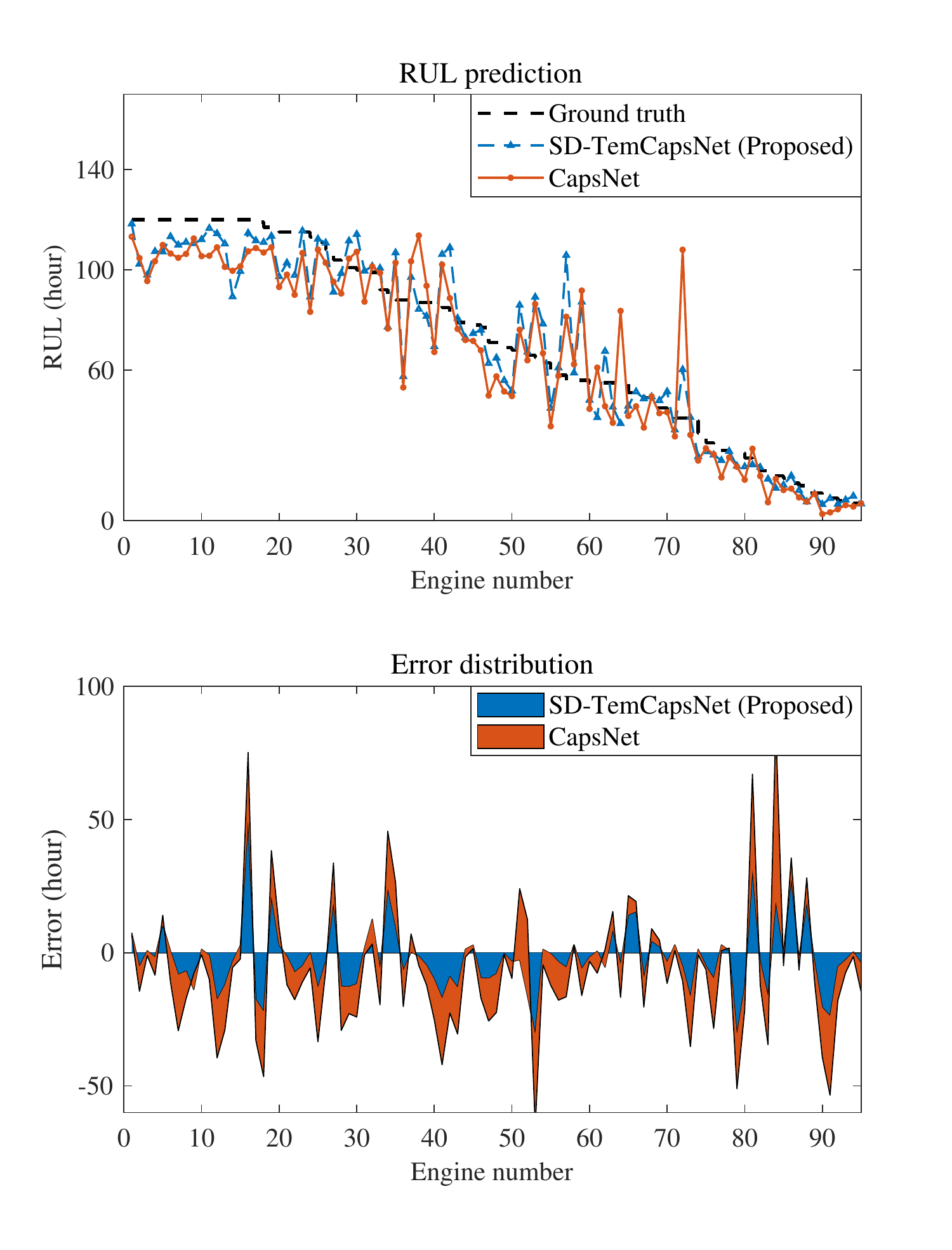}
\end{minipage}
}
\subfigure[]
{
\begin{minipage}[t]{0.45\linewidth}
\centering
\includegraphics[width=6.5cm]{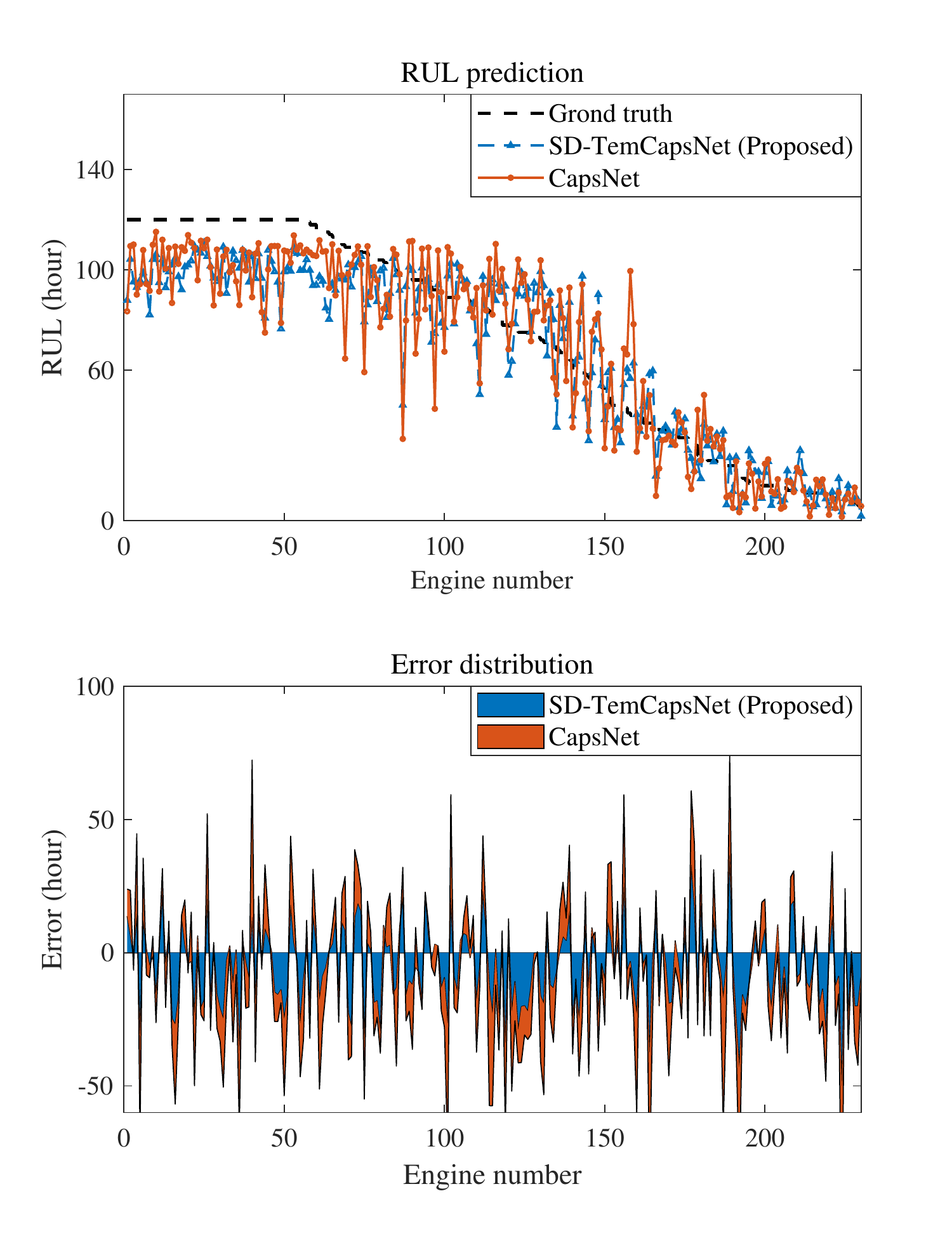}
\end{minipage}
}
\centering
\caption{Prediction performance comparison between SD-TemCapsNet and CapsNet on testing datasets for (a) FD001, (b) FD002, (c) FD003, and (d) FD004.}
\end{figure}

\begin{table}[H]
\renewcommand{\arraystretch}{1.2}
\centering
\scriptsize
\caption{Ablation study of temporal dynamics and slow-varying dynamics for aircraft engine system.}
\begin{tabular}{c c c c c c c}
\hline
\hline
\multirow{2}{*}{\textbf{Dataset}} & \multirow{2}{*}{\textbf{Metrics}}  & \multirow{2}{*}{\textbf{CapsNet [12]}} & \multicolumn{2}{c}{\textbf{TemCapsNet}} & \multicolumn{2}{c}{\textbf{SD-CapsNet}} \\
\cline{4-7}
 &  &  & Accuracy & IMA &  Accuracy & IMA \\
\hline
\multirow{2}{*}{FD001} & $RMSE$ & 12.58±0.25 & 11.89±0.14 & 5.48$\%$ & 11.92±0.24 & 5.25$\%$ \\
& $SF$ & 276.34±25.95 & 202.65±8.57 & 26.67$\%$ & 182.82±12.22 & 33.84$\%$ \\
\hline
\multirow{2}{*}{FD002} & $RMSE$ & 16.30±0.23 & 14.93±0.27 & 8.40$\%$ & 15.28±0.44 & 6.26$\%$\\
& $SF$ & 1229.72±53.07 & 601.88±54.69 & 51.06$\%$ & 611.02±23.16  & 50.36$\%$\\
\hline
\multirow{2}{*}{FD003} & $RMSE$  & 11.71±0.26 & 12.54±0.25 & -7.09$\%$ & 11.75±0.31 & -0.34$\%$ \\
& $SF$ & 283.81±29.46 & 202.14±10.58 & 28.78$\%$ & 191.70±38.28 & 32.45$\%$ \\
\hline
\multirow{2}{*}{FD004} & $RMSE$  & 18.96±0.27 & 17.63±0.22 & 7.01$\%$ & 17.03±0.30  & 10.18$\%$ \\
& $SF$ & 2625.64±266.83 & 1473.38±136.16 & 43.88$\%$ & 874.14±53.43 & 66.71$\%$\\
\hline
\hline
\end{tabular}
\begin{tablenotes}
\item[1] IMA is short for improved prediction accuracy regarding $RMSE$ compared to CapsNet.
\end{tablenotes}
\end{table}

\subsubsection{Ablation Studies} An ablation analysis has been carried out to evaluate the individual contribution of slow-varying dynamics and temporal dynamics through removing one of them from SD-TemCapsNet in turn. Taking the original CapsNet as the baseline, Table V presents the estimation errors of TemCapsNet and SD-CapsNet. Besides, comparisons have been made with the original CapsNet.

\begin{itemize}
\item {\textit{Contribution of slow-varying dynamics:}} SD-CapsNet evaluates the importance of slow-varying dynamics, which excludes the temporal attention ability with LSTM discussed in Section III. B(4). As seen in Table V, the slow-varying dynamics accounts for outstanding error reduction compared to the traditional CapsNet. Specifically, for $RMSE$, estimation error has been reduced by 5.25$\%$, 6.26$\%$, and 10.18$\%$ for FD001, FD002, and FD004, respectively. Moreover, estimation errors have been reduced by at least $30\%$ for all datasets concerning the index $SF$. The introduction of slow features could simultaneously reduce the estimation error of $RMSE$ and $SF$ for all cases. Therefore, it proves that the utilization of slow-varying dynamics enhances the precision of the results.

\item {\textit{Contribution of temporal dynamics:}} TemCapsNet assesses the utility of temporal dynamics, replacing $\mathbf T_c$ with $\mathbf X_{d,c}$ and keeping the remaining procedure of SD-TemCapsNet same. In Table V, adding temporal attention ability with a LSTM decreases the estimation error evaluated by both $RMSE$ and $SF$. Especially, the index $SF$ has been reduced by at least $20\%$ for all cases.
\end{itemize}

Compared to temporal dynamics, slow-varying dynamics contributes to more performance boost evaluated by the metrics $SF$. Moreover, integrating slow-varying and temporal dynamics yields more accurate results than the individual one.

All experiments are fairly conducted and compared under the same computing ability, which is achieved by a workstation equipped with eight processors of Intel Xeon E5-2620 v4 and a multi-graphics processor unit of NVIDIA GeForce GTX 1080Ti. The computing complexity of the proposed method about training time and the number of parameters are compared with available counterparts, as shown in Fig. 12. The amount of parameters of SD-TemCapsNet has increased a lot in comparison with other methods except for FD003, in which the small number of filters contributes to reducing the number of parameters. However, the training time does not increase too much, especially for FD001 and FD004. That is, the proposed method needs fewer iteration epochs, and the large $L$ will further reduce the amount of training data.

\subsection{RUL Estimation for Degraded Milling Machine}
This section verifies the performance of the proposed SD-TemCapsNet on a milling manufacturing process with cyclic-degradation behaviour. Fig. 14 exhibits the experiment platform \cite{Ref28}, which senses six measurements under three extra operation parameters. For model development and verification purpose, 166 runs operated at 15 testing cases are collected, in which 109 runs for material Type 1 (cast iron) and 57 runs for material Type 2 (steel). The number of runs for each testing case depends on the degree of flank wear between runs. For the milling manufacturing procedure, the cutter needs to be replaced when the flank wear reaches the preset threshold. Typically, the labeled value of RUL is set to be 0 when the flank wear is more than 0.45 for the first time, according to expert knowledge \cite{Ref15}. Taking the flank wear collected in the first operating condition as an instance, the yielded piece-wise ground truth of flank wear and the corresponding labels for estimation are shown in Fig. 15.

\begin{figure}[H]
\centering
\includegraphics[scale=0.4]{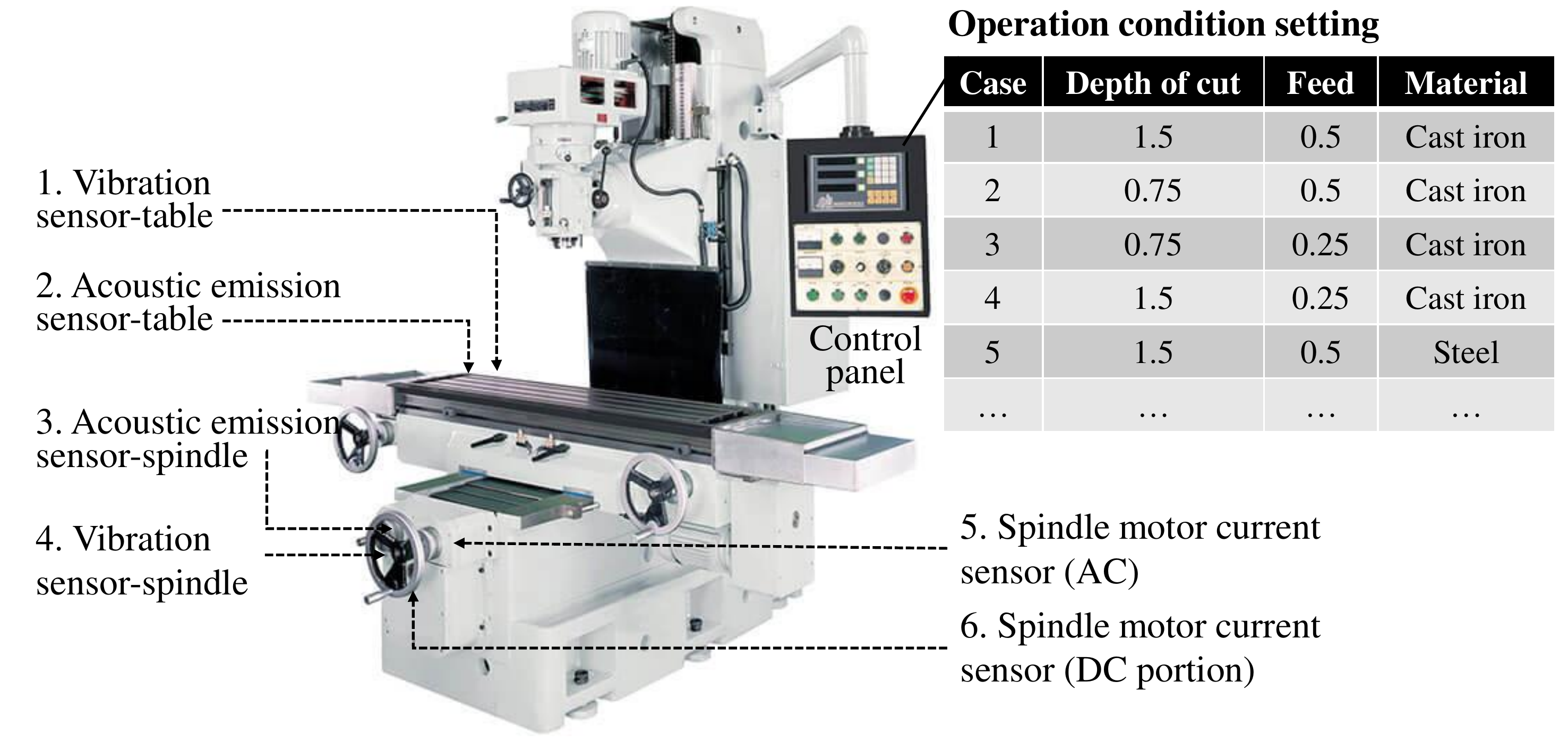} \\
\begin{center}
{\caption{The overall structure of milling machine with deployed sensors.}}
\end{center}
\label{FIG14}
\end{figure}

\begin{figure}[!ht]
\centering
\includegraphics[scale=0.6]{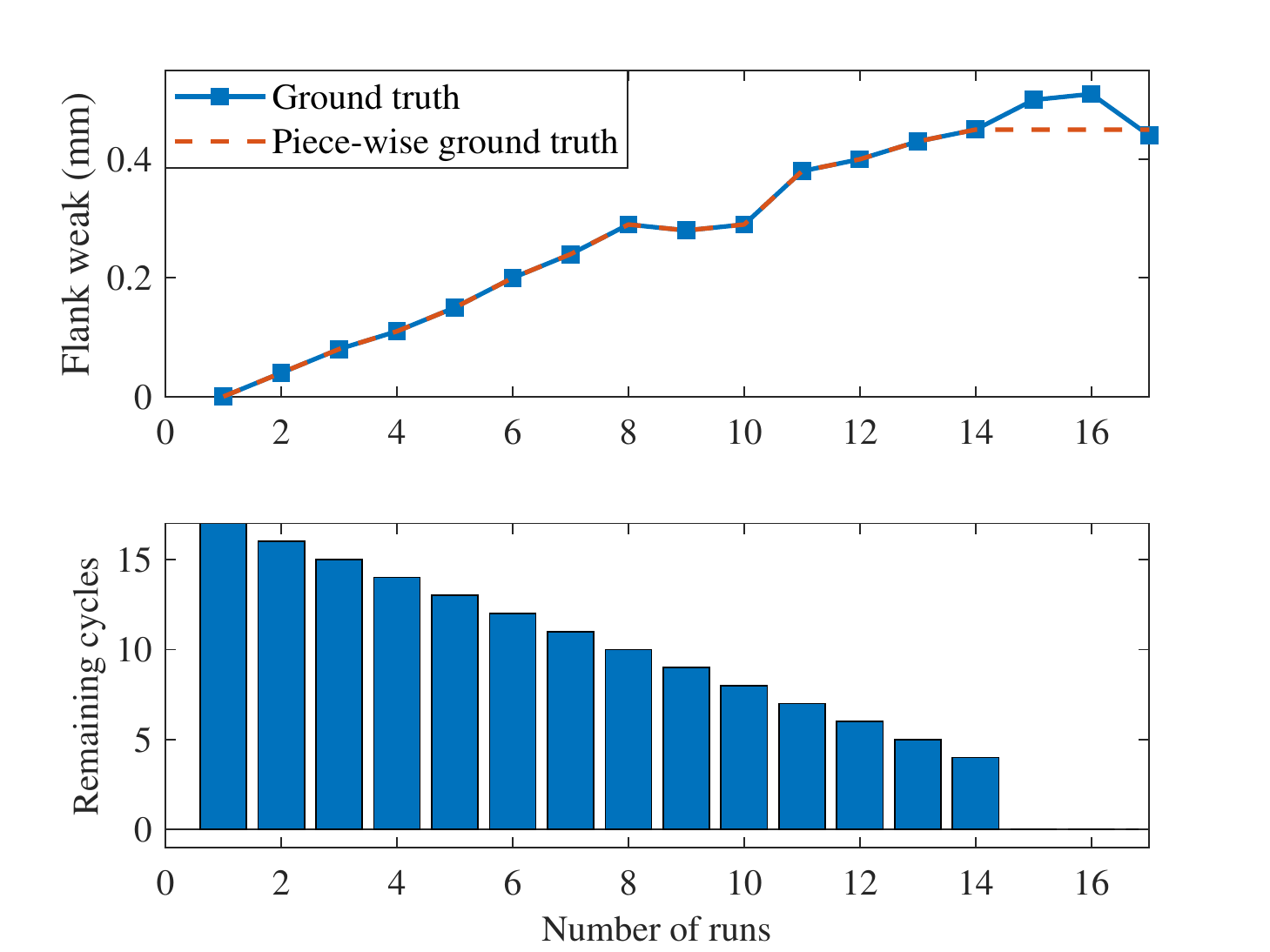} \\
\begin{center}
\caption{The ground truth and its piece-wise function of flank wear of the first operating condition in milling dataset.}
\end{center}
\label{FIG15}
\end{figure}

\subsubsection{Slow Feature Representation} Then slow-varying space derived from the normal runs encompasses the first five eigenvectors according to Eq. (4). And the remaining one is treated as residual space. Fig. 16 visualizes the obtained slow features by taking all the runs from the first operating case described in Fig. 14 as an example. It is worth noting that each run lasts 90 samples through down-sampling, yielding each grid of Fig. 16 corresponds to a cycle. Fig. 16(a) presents the varying trend of slow features over runs, reflecting continuous wear of the cutter. Contrarily, the residual feature plotted in Fig. 16(b) almost keeps the same distribution, indicating the insensitivity to degradation. By performing the same feature extraction for both training and testing data, the hybrid features are ready for modelling.

\begin{figure}[H]
\subfigure[]
{
\begin{minipage}[t]{1\linewidth}
\centering
\includegraphics[width=9cm]{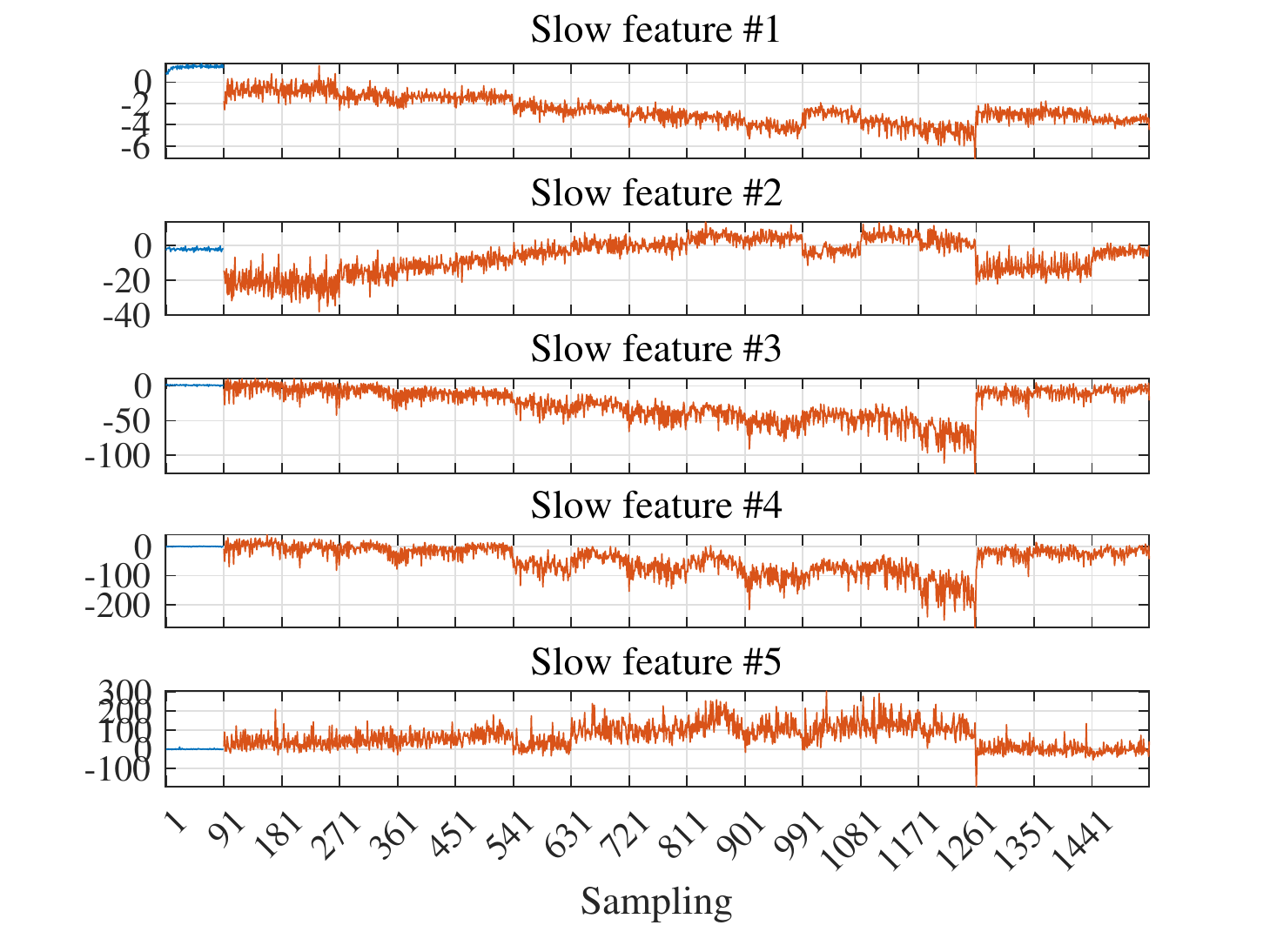}
\end{minipage}
}

\subfigure[]
{
\begin{minipage}[t]{1\linewidth}
\centering
\includegraphics[width=9cm]{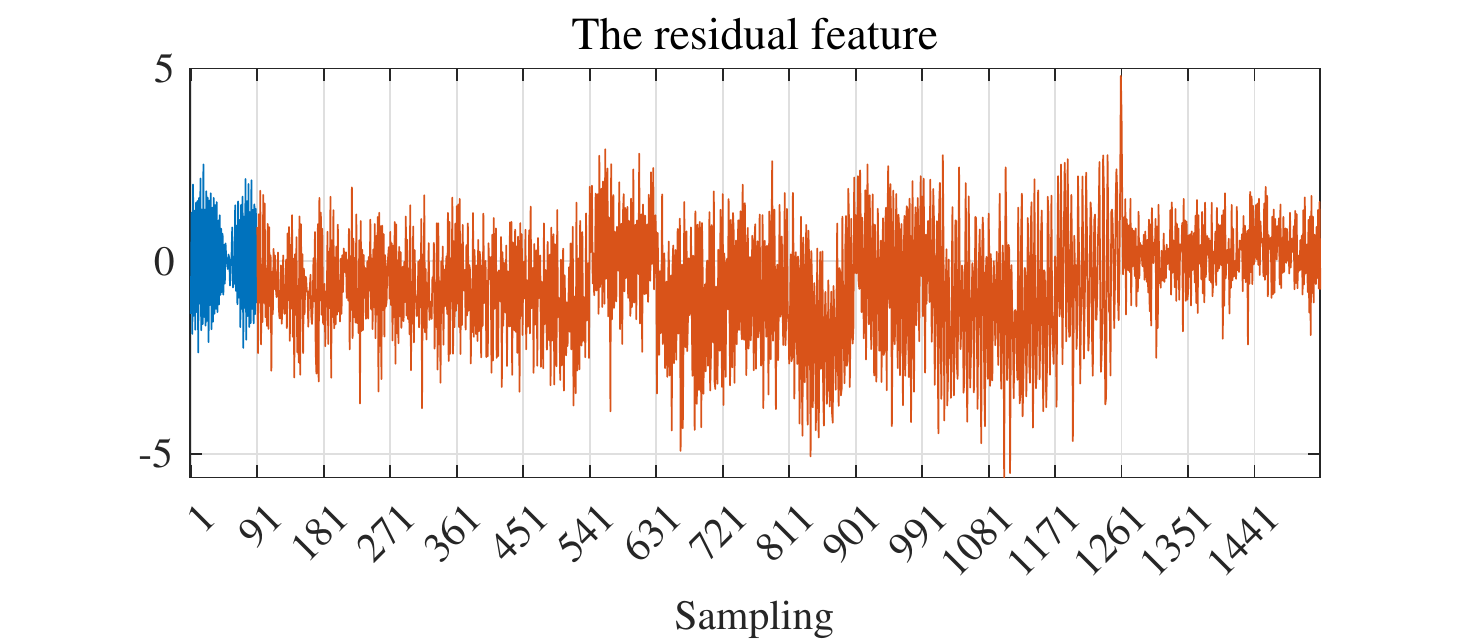}
\end{minipage}
}
\begin{center}
\caption{The extracted (a) slow features and (b) the residual feature from the first operating condition in training data (The blue line indicates the normal stage and the orange line stands for the fault stage).}
\end{center}
\label{FIG16}
\end{figure}

The same training data and testing data are used for a fair comparison with the mainstream methods \cite{Ref15}. The first 9 cases of material Type 1 and the first 2 cases of material Type 2 serve as training samples. The remaining 4 cases are employed as testing samples.

\subsubsection{Piece-wise Data Normalization} We pick up the first run of each operating condition in training data to generate the normalization information, as they behave normally with the zero flank wear. Then the remaining data are normalized according to the way given in Eq. (2).

\subsubsection{Network Configuration} For the proposed SD-TemCapsNet, the specific parameters of three layers are listed in Table VI. CNN and LSTM-based estimation models adopt the reported network structure \cite{Ref15} for comparison. For the original CapsNet, we remove the LSTM layer from Table VI for easy comparison.

\begin{table}[H]
\renewcommand{\arraystretch}{1.2}
\centering
\caption{Configurations of SD-TemCapsNet for milling machine.}
\scriptsize
\begin{tabular}{c c c}
\hline
\hline
\multirow{1}{*}{\textbf{Layer Name}} & \multirow{1}{*}{\textbf{Parameter Name}}  & \multirow{1}{*}{\textbf{Parameter Value}}\\
\hline
\multirow{2}{*}{Input} & {Epoch}  & 80 \\
\cline{2-3}
& {Window length}  & 20 \\
\hline
\multirow{3}{*}{Convolution} & {Filters}  & 24  \\
\cline{2-3}
& {Kernal size}  & (1,3) \\
\cline{2-3}
& {Strides}  & (1,3) \\
\hline
\multirow{4}{*}{Basic capsule} & {Dimensions}  & 3 \\
\cline{2-3}
& {Channels} & 8 \\
\cline{2-3}
& {Kernal size} & (1,3) \\
\cline{2-3}
& {Strides}  & (1,3) \\
\hline
\multirow{2}{*}{Advanced capsule} & {No. of Capsules}  & 5 \\
\cline{2-3}
& {Dimensions}  & 6 \\
\hline
{LSTM} & {Units}  & 16 \\
\hline
\multirow{3}{*}{Output} & {The first FNN} & {No. of Neurons=200, Dropout ratio=0.2} \\
& {The second FNN}  & {No. Neurons=100, Dropout ratio=0.2} \\
& {The third FNN}     & {No. Neurons=1} \\
\hline
\hline
\end{tabular}
\end{table}

\subsubsection{Performance Comparison} With the configured network, SD-TemCapsNet yields the best results compared to other approaches, as shown in Table VII. Compared to the original CapsNet, the estimation error has been reduced by 19.54$\%$. Moreover, Fig. 17 gives detailed prediction results.

\begin{table}[H]
\scriptsize
\renewcommand{\arraystretch}{1.2}
\centering
\caption{Performance comparison between the the proposed method and its counterparts for degraded milling machine system regarding $RMSE$.}
\begin{tabular}{c c}
\hline
\hline
\multirow{1}*{\textbf{Model Name}} & \multicolumn{1}{c}{\textbf{Performance index ($RMSE$)}}\\
\hline
CNN [15] & 6.16   \\
\hline
LSTM [15] & 2.80 \\
\hline
CapsNet & 2.66  \\
\hline
SD-TemCapsNet (Proposed) & 2.14 \\
\hline
IMA compared to CapsNet & 19.54$\%$ \\
\hline
\hline
\end{tabular}
\begin{tablenotes}
\item[1] IMA is short for improved prediction accuracy regarding $RMSE$.
\end{tablenotes}
\end{table}

\begin{figure}[H]
\centering
\includegraphics[scale=0.7]{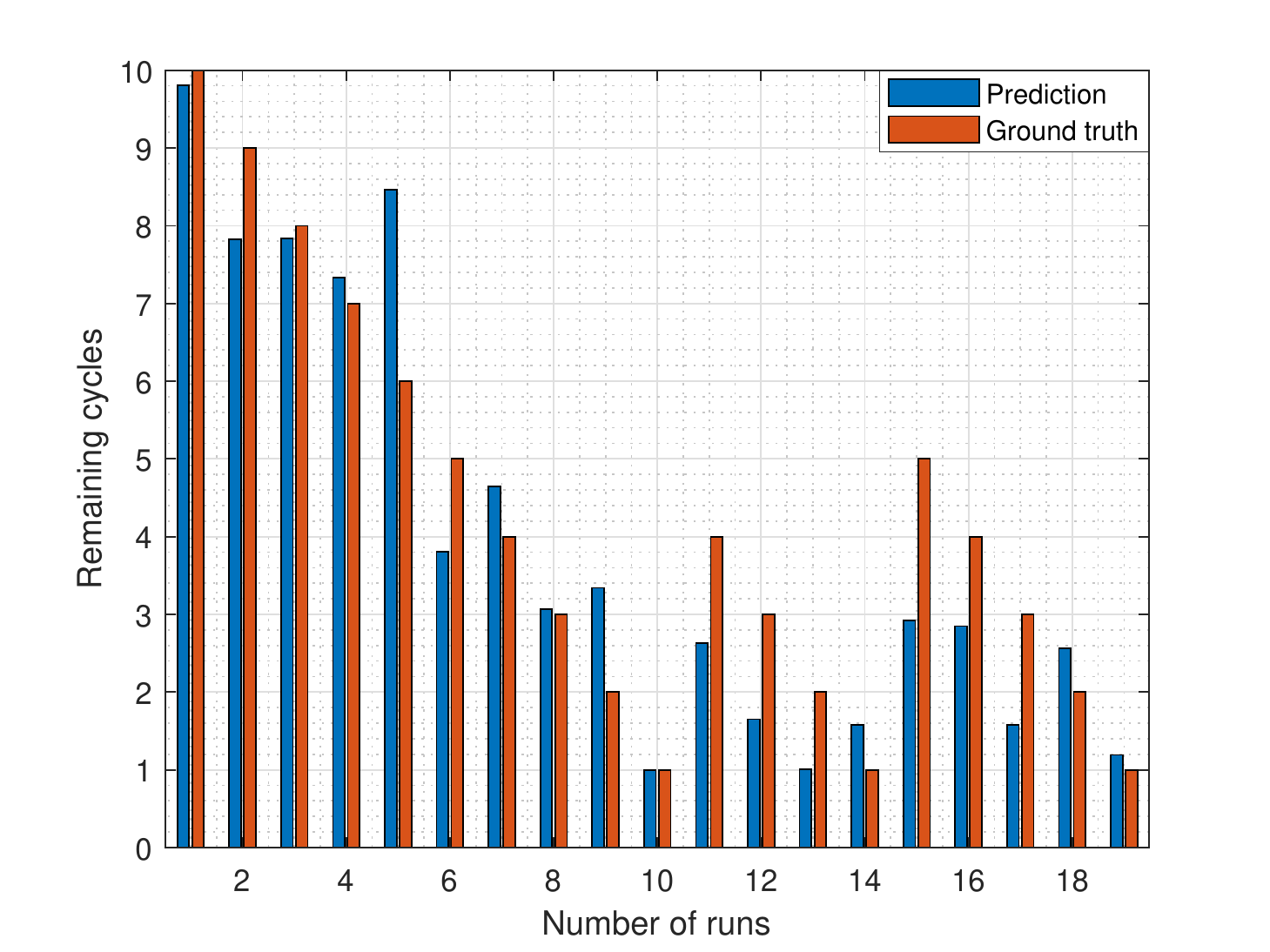} \\
\begin{center}
{\caption{Prediction performance of SD-TemCapsNet for milling dataset.}}
\end{center}
\label{Fig17}
\end{figure}

\subsubsection{Ablation Studies}
To further evaluate the significance of slow-varying dynamics and temporal dynamics in the milling dataset, SD-CapsNet and TemCapsNet are configured utilizing the similar ways reported in the aircraft engine case to conduct ablation analysis. The original CapsNet is employed as a baseline, which gains the estimation error of 2.66 with $RMSE$. Table VIII systemically summarizes the estimation results of SD-CapsNet and TemCapsNet, both of which reduce the estimation error by around 10$\%$ compared to the baseline model. In all, the introduction of both dynamics contributes to a performance enhancement.

\begin{table}[H]
\scriptsize
\renewcommand{\arraystretch}{1.2}
\centering
\caption{Ablation study of temporal dynamics and slow-varying dynamics for degraded milling machine system.}
\begin{tabular}{c c c c}
\hline
\hline
\multirow{2}*{\textbf{Index}} & \multicolumn{3}{c}{\textbf{Model Name}} \\
\cline{2-4}
& \multicolumn{1}{c}{\textbf{CapsNet}} & \multirow{1}*{\textbf{TemCapsNet}} & \multicolumn{1}{c}{\textbf{SD-CapsNet}} \\
\hline
$RMSE$ & 2.66 & 2.42 & 2.36  \\
\hline
IMA compared to CapsNet & --  & 9.02$\%$ & 11.28$\%$ \\
\hline
\hline
\end{tabular}
\begin{tablenotes}
\item[1] IMA is short for improved prediction accuracy regarding $RMSE$.
\end{tablenotes}
\end{table}

\section{Conclusion and Future work}
Deep neural networks play significant roles in remaining useful life estimation for modern machinery processes in the era of big data and artificial intelligence. Research is being actively conducted towards further improving them not only by advanced intelligent algorithms but also via in-depth process understanding. Motivated by the success of capsule network (CapsNet) over convolutional neural networks (CNN) at image-based inference tasks, we presented the Slow-varying Dynamics assisted CapsNet (SD-TemCapsNet). This novel approach incorporates both CapsNet and long-short-term memory for run-to-failure time-series learning. We have verified that SD-TemCapsNet outperforms present methods from both the overall performance indices and sectional error distribution with two experimental cases. The above results indicate that CapsNet is a competing alternative to CNNs for designing inference modules in future RUL estimation models.

Although some topics are beyond the scope of this work, they may deserve attention to enhance RUL estimation performance, leaving the following directions for consideration:
\begin{itemize}
\item It could be interesting to put forward an automatic way to locate the change point of the piece-wise function, where the machinery performance begins to degrade. This would aid in releasing the requirement on expert knowledge and further boost RUL estimation performance.

\item It is impractical to assume that data distributions hold consistency for machines under varying operating conditions. As such, transfer learning may provide a feasible solution to leverage existing source models or aprior knowledge to new operating conditions.
\end{itemize}

\ifCLASSOPTIONcaptionsoff
  \newpage
\fi

\end{document}